\documentclass{article} 
\usepackage[T1]{fontenc}
\PassOptionsToPackage{table}{xcolor}
\usepackage{iclr2027_conference,times}

\usepackage{amsmath,amsfonts,bm}

\def\eqref#1{equation~\ref{#1}}

\def\1{\bm{1}}

\DeclareMathAlphabet{\mathsfit}{\encodingdefault}{\sfdefault}{m}{sl}
\SetMathAlphabet{\mathsfit}{bold}{\encodingdefault}{\sfdefault}{bx}{n}

\renewcommand{\eqref}[1]{\textup{(\ref{#1})}}

\usepackage[hidelinks]{hyperref}
\usepackage{url}
\usepackage{booktabs,multirow,graphicx}
\usepackage{wrapfig}
\usepackage{tikz}
\usepackage{pgfplots}
\pgfplotsset{compat=1.17}
\usetikzlibrary{shapes.geometric}
\usepackage{subcaption}
\usetikzlibrary{arrows.meta,calc,fit,positioning,shapes.geometric}
\usepackage{xcolor}

\newcounter{algorithm}
\newenvironment{paperalgorithm}[1]{%
  \refstepcounter{algorithm}\par\medskip\noindent
  \begin{minipage}{\linewidth}\hrule\vspace{4pt}
  \textbf{Algorithm \thealgorithm: #1}\vspace{3pt}\hrule\vspace{4pt}\small
}{%
  \vspace{3pt}\hrule\end{minipage}\par\medskip
}

\newcommand{\name}{RingMatch}

\title{Beyond Local Refinement: Wide-Area Precise Cross-View Localization}

\newcommand{\projectpageurl}{https://qghuxmu.github.io/RingMatch/}

\author{
Qingguo Hu$^{1}$, Xinyu Ma$^{3}$, Yujie Lin$^{1}$, Yaoxiang Wang$^{1}$, Qi Shi$^{2}$, Zonghao Guo$^{2}$,\\[2pt]
\bfseries Chi Chen$^{2,*}$, Jinsong Su$^{1,}$\thanks{Corresponding authors.}\\[4pt]
{\normalfont $^{1}$Xiamen University\quad $^{2}$Tsinghua University\quad $^{3}$University of Macau}\\[2pt]
{\normalfont\small\href{mailto:huqingguo@stu.xmu.edu.cn}{\nolinkurl{huqingguo@stu.xmu.edu.cn}}\quad
\href{mailto:jssu@xmu.edu.cn}{\nolinkurl{jssu@xmu.edu.cn}}}\\[5pt]
{\normalfont\projectpageline}
}

\definecolor{projecturl}{rgb}{0.4,0.19609,0.49805}

\newcommand{\emptyprojectpageurl}{}
\newcommand{\projectpageline}{%
  {\small\color{black}%
  \ifx\projectpageurl\emptyprojectpageurl
    Project Page:%
  \else
    \href{\projectpageurl}{Project Page:\enspace
      {\color{projecturl}\nolinkurl{\projectpageurl}}}%
  \fi}%
}

\iclrfinalcopy 
\begin{document}

\maketitle
\lhead{Preprint.}
\renewcommand{\headrulewidth}{0.4pt}

\begin{abstract}
Fine-grained cross-view localization aims to precisely localize a ground camera against geo-referenced aerial imagery, but current methods typically assume a narrow geographic prior, substantially limiting their applicability when upstream localization remains coarse. To bridge this gap, we study this task over much larger aerial regions. However, this seemingly simple extension fundamentally changes the localization problem: the growing disparity in spatial extent between ground and aerial views makes reliable cross-view matching difficult to establish directly, while the increasing number of look-alike locations in larger regions introduces substantial visual ambiguity.
To address these challenges, we propose \name{}, a search-and-verification framework for wide-area cross-view localization. \name{} first leverages predictive uncertainty to uncover plausible locations and progressively zooms into their surroundings to obtain increasingly fine-grained evidence, yielding a compact set of promising candidates. We then introduce the \emph{depth ring}, a novel geometric representation that reveals subtle variations in the spatial layout around a location and serves as a view-shared spatial signature for an LVLM verifier to resolve residual ambiguity.
To benchmark this setting, we construct VIGOR-Pro by composing VIGOR aerial tiles into contiguous regions with spatial extents of up to \(400\,\mathrm{m}\). Experiments show that \name{} maintains strong localization performance as the aerial region expands, consistently outperforming representative end-to-end and retrieval-based methods.
\end{abstract}

\section{Introduction}

Determining where an image was captured from visual observations is a fundamental problem with broad applications in navigation, robotics, and public safety.
At the global scale, image geo-localization provides a coarse location estimate for a ground image from geographic cues in the scene~\citep{vivanco2023geoclip,li2025globe}.
To achieve finer localization, aerial imagery has been widely used as a reference owing to its broad coverage and rich scene details~\citep{deuser2023sample4geo}.
Within this paradigm, ongoing research has increasingly focused on the final stage of the overall pipeline, known as fine-grained cross-view localization~\citep{zhu2021vigor}, which seeks to estimate the ground camera's precise metric position within a geo-referenced aerial image. By leveraging advances in cross-view representation learning~\citep{xia2022metric,fervers2023uncertainty,xia2024ccvpe} and fine-grained spatial matching~\citep{wang2023fine,lentsch2023slicematch,xia2025fg2}, existing approaches have achieved notable improvements in metric localization accuracy.


Despite these advances, a critical gap remains before existing fine-grained cross-view localization methods can be reliably deployed in real-world applications. Specifically, current methods largely operate in a \emph{local-refinement} regime, assuming that the camera has been narrowed down to a tightly bounded aerial region before estimating its precise position.
On VIGOR~\citep{zhu2021vigor}, a widely used benchmark, the commonly adopted setting restricts the camera to a prior area of only about \(35\times35\) meters~\citep{lentsch2023slicematch,xia2024ccvpe,xia2025fg2}. Such a narrow range requires the upstream localization system to reduce its uncertainty to a comparable scale. In practice, however, upstream localization from GPS/GNSS or image retrieval may remain coarse, and such a precise prior may not be available~\citep{xia2022metric,shi2023boosting}. This mismatch limits the integration of cross-view localization into practical visual localization systems.


A natural way to accommodate coarser upstream localization priors is to enlarge the aerial region. However, this seemingly straightforward extension fundamentally changes the localization problem.
As the aerial region expands, the ground observation covers only a small area within it, making it difficult to directly establish reliable cross-view matching. As shown in Figure~\ref{fig:challenges:degradation}, expanding the aerial region can severely degrade existing methods.
This growing spatial-extent disparity inherently shifts the task from local refinement to wide-area spatial search.
Meanwhile, larger aerial regions often contain more recurring structures, creating multiple look-alike locations compatible with the same ground observation. Such locations often share similar appearance but differ in subtle details of their spatial layouts (see Figure~\ref{fig:challenges:aliasing}).
Resolving this ambiguity requires recognizing and comparing these spatial cues across large viewpoint disparities, yet such fine-grained cross-view spatial reasoning remains challenging for current multimodal models~\citep{jeong2026xvr,liu2026cvsbench}.

%
%
%
\ifdefined\challengefigshared \fi
\newcommand{\challengefigshared}{}

\definecolor{cHCNet}{HTML}{D55E00}   
\definecolor{cFG2}  {HTML}{0072B2}   
\definecolor{cCCVPE}{HTML}{009E73}   
\definecolor{cGT}   {HTML}{D66A14}   
\definecolor{cLA}   {HTML}{1F77B4}   
\definecolor{cLB}   {HTML}{D66EB4}   
\definecolor{cGround}{HTML}{3D3D3D}  

\tikzset{
    aerialbox/.style={line width=1.0pt},
    croptag/.style={font=\tiny\bfseries, text=white, inner sep=1.4pt},
    groundtag/.style={font=\tiny, text=white, inner sep=1.1pt},
}

\newcommand{\ptcam}[3]{%
    \node[star, star points=5, star point ratio=2.2, fill=cGT, draw=white,
          line width=0.6pt, minimum size=#3, inner sep=0pt] at (#1,#2) {};}
\newcommand{\ptcand}[4]{%
    \node[circle, fill=#3, draw=white, line width=0.6pt, minimum size=#4,
          inner sep=0pt] at (#1,#2) {};}

\newcommand{\scalebarat}[4]{%
    \fill[black, opacity=0.30] (#1-0.10,#2-0.10) rectangle (#1+#3+0.10,#2+0.28);
    \draw[white, line width=1.0pt] (#1,#2) -- (#1+#3,#2);
    \draw[white, line width=1.0pt] (#1,#2-0.06) -- (#1,#2+0.06);
    \draw[white, line width=1.0pt] (#1+#3,#2-0.06) -- (#1+#3,#2+0.06);
    \node[font=\tiny, text=white, anchor=south] at (#1+#3/2,#2-0.05) {#4};}

\newcommand{\groundinset}[1]{%
    \def\giw{2.00}\def\gih{1.00}
    \def\gix{0.10}\def\giy{0.10}
    \pgfmathsetmacro{\gixx}{\gix+\giw}%
    \pgfmathsetmacro{\giyy}{\giy+\gih}%
    \fill[black, opacity=0.18] (\gix+0.04,\giy-0.04)
                    rectangle (\gixx+0.04,\giyy-0.04);
    \node[anchor=south west, inner sep=0pt] at (\gix,\giy)
         {\includegraphics[width=\giw cm, height=\gih cm]{#1/panorama.jpg}};
    \draw[white,     line width=1.0pt] (\gix,\giy) rectangle (\gixx,\giyy);
    \draw[black!45,  line width=0.4pt] (\gix-0.018,\giy-0.018)
                             rectangle (\gixx+0.018,\giyy+0.018);
          }

\newcommand{\degradationpanel}{%
\begin{tikzpicture}
\hspace*{-0.2cm}
\begin{axis}[
    width=6.4cm, height=5.4cm,
    xmin=0, xmax=435,
    ymode=log, log basis y=10,
    ymin=1.3, ymax=120,                 
    xtick={35,100,200,300,400},
    ytick={2,5,10,20,50,100},              
    yticklabels={2,5,10,20,50,100},        
    minor ytick={3,4,6,7,8,9,30,40,60,70,80,90,100},
    xlabel={Aerial Extent (m)},
    ylabel={Mean Distance Error (m) $\downarrow$},
    y label style={at={(axis description cs:-0.125,0.5)},anchor=south},
    label style={font=\small},
    tick label style={font=\small},
    grid=major, ymajorgrids, yminorgrids,
    major grid style={line width=0.5pt, draw=gray!30},
    minor grid style={line width=0.4pt, draw=gray!15},
    legend style={font=\small, draw=gray!45, fill=white, inner sep=3pt},
    legend pos=south east,             
    legend cell align=left,
]
\addplot[cHCNet, line width=1.4pt, mark=o, mark size=2.6pt]
    coordinates {(35,2.65) (100,10.52) (200,29.54) (400, 85.62)};
\addlegendentry{HC-Net}
\addplot[cCCVPE, line width=1.4pt, mark=o, mark size=2.6pt]
    coordinates {(35,3.60) (100,9.63) (200,14.07) (400, 27.90)};
\addlegendentry{CCVPE}
\addplot[cFG2, line width=1.4pt, mark=o, mark size=2.6pt]
    coordinates {(35,1.95) (100,9.76) (200,18.38) (400, 38.41)};
\addlegendentry{FG$^2$}
\end{axis}
\end{tikzpicture}
}

\newcommand{\cropcolumn}[1]{%
    \def\satcropgap{0.25}%
    \pgfmathsetmacro{\cropx}{5.00+\satcropgap}%
    \pgfmathsetmacro{\cropright}{\cropx+1.45}%
    \pgfmathsetmacro{\cropcenter}{\cropx+0.725}%
    \node[anchor=south west, inner sep=0pt] at (\cropx,3.55)
         {\includegraphics[width=1.45cm, height=1.45cm]{#1/crop_gt.jpg}};
    \draw[aerialbox, cGT] (\cropx,3.55) rectangle (\cropright,5.00);
    \node[croptag, fill=cGT, anchor=north west] at (\cropx,5.00) {GT};
    \ptcam{\cropcenter}{4.275}{6pt}
    \node[anchor=south west, inner sep=0pt] at (\cropx,1.775)
         {\includegraphics[width=1.45cm, height=1.45cm]{#1/crop_l1.jpg}};
    \draw[aerialbox, cLA] (\cropx,1.775) rectangle (\cropright,3.225);
    \node[croptag, fill=cLA, anchor=north west] at (\cropx,3.225) {A};
    \ptcand{\cropcenter}{2.500}{cLA}{4.4pt}
    \node[anchor=south west, inner sep=0pt] at (\cropx,0)
         {\includegraphics[width=1.45cm, height=1.45cm]{#1/crop_l2.jpg}};
    \draw[aerialbox, cLB] (\cropx,0) rectangle (\cropright,1.45);
    \node[croptag, fill=cLB, anchor=north west] at (\cropx,1.45) {B};
    \ptcand{\cropcenter}{0.725}{cLB}{4.4pt}}

\newcommand{\subcapdegradation}{Degradation over larger aerial regions.}
\newcommand{\subcapaliasing}{Ambiguity from look-alike locations.}

\begin{figure}[t]
\centering
\begin{subfigure}[t]{0.49\textwidth}
\vspace{0pt}
\centering
\scalebox{0.96}{\degradationpanel}
\caption{\subcapdegradation}
\label{fig:challenges:degradation}
\end{subfigure}%
\hspace{-0.0\textwidth}%
\begin{subfigure}[t]{0.49\textwidth}
\vspace*{0.03cm}
\centering
\scalebox{0.90}{%
\begin{tikzpicture}[scale=0.96, transform shape]
\node[anchor=south west, inner sep=0pt] at (0,0)
     {\includegraphics[width=5cm, height=5cm]{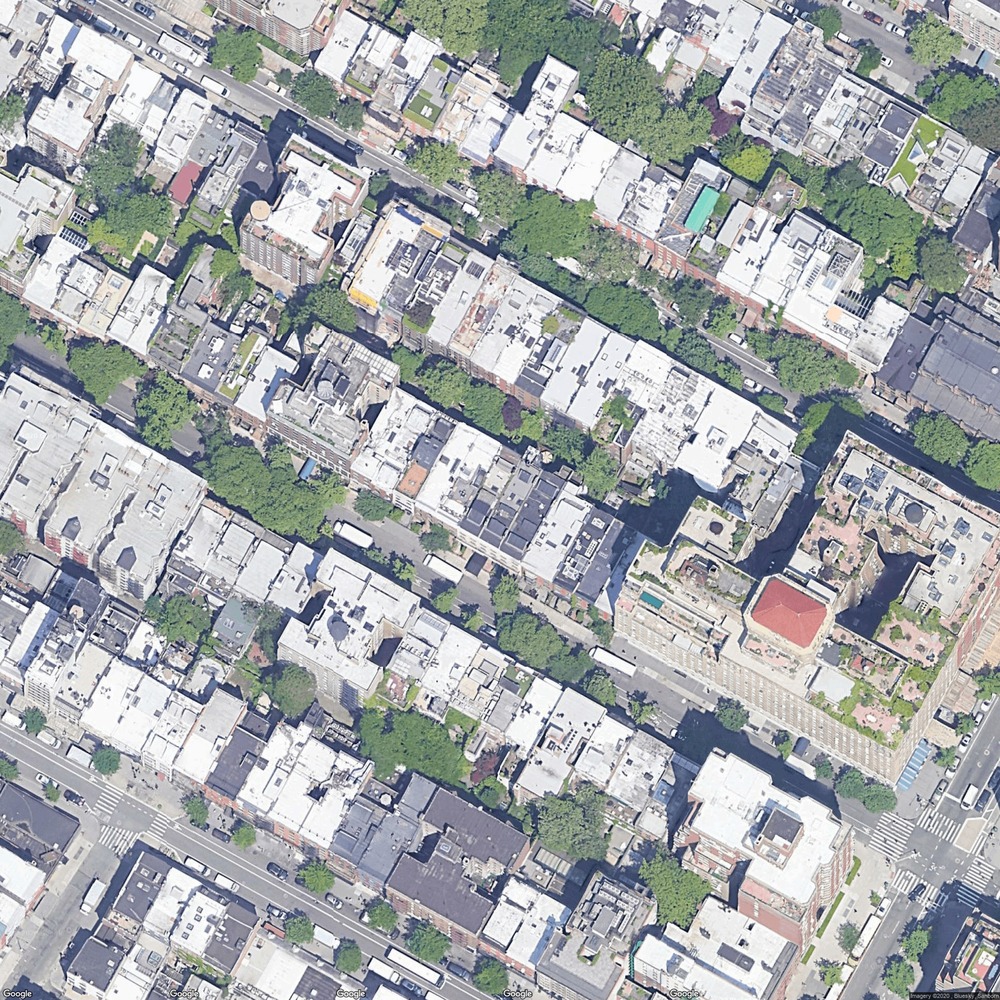}};
\draw[aerialbox, cGT] (3.621,2.517) rectangle (4.540,3.436);
\draw[aerialbox, cLA] (1.389,1.841) rectangle (2.308,2.761);
\draw[aerialbox, cLB] (0.905,4.028) rectangle (1.825,4.948);
\ptcand{1.848}{2.301}{cLA}{4.4pt}
\ptcand{1.365}{4.488}{cLB}{4.4pt}
\ptcam{4.081}{2.976}{7pt}
\scalebarat{3.66}{0.16}{1.150}{50\,m}
\groundinset{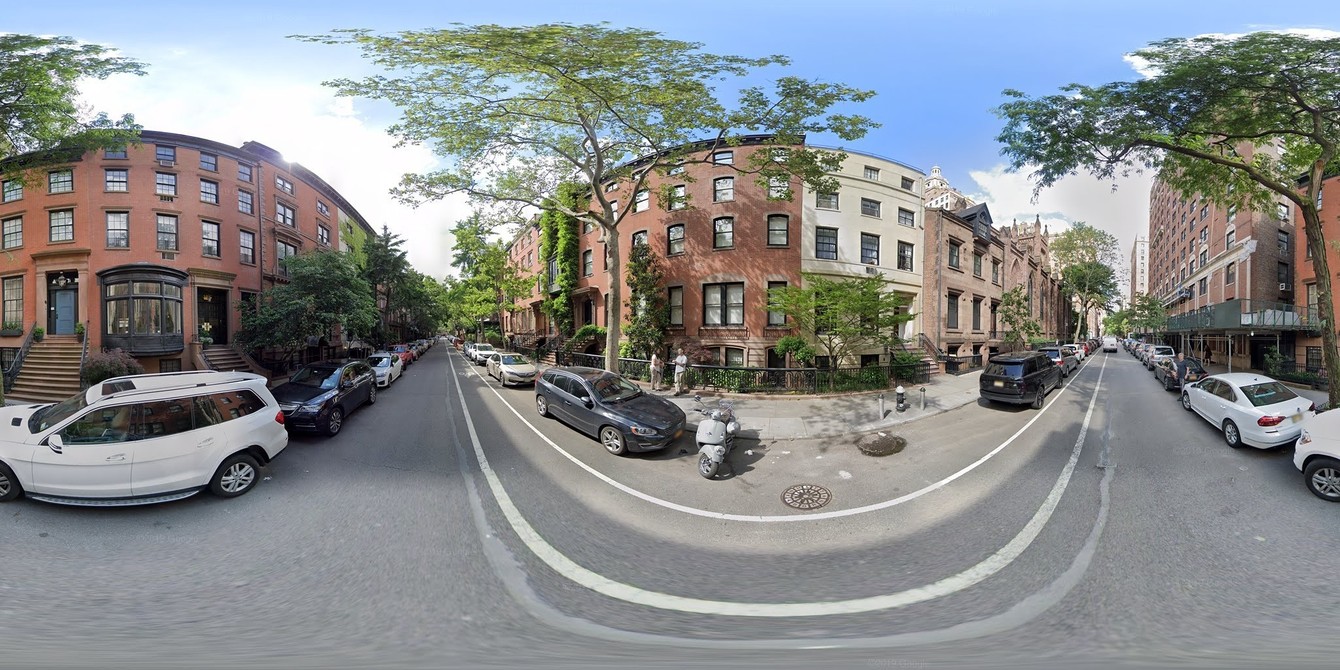}
\cropcolumn{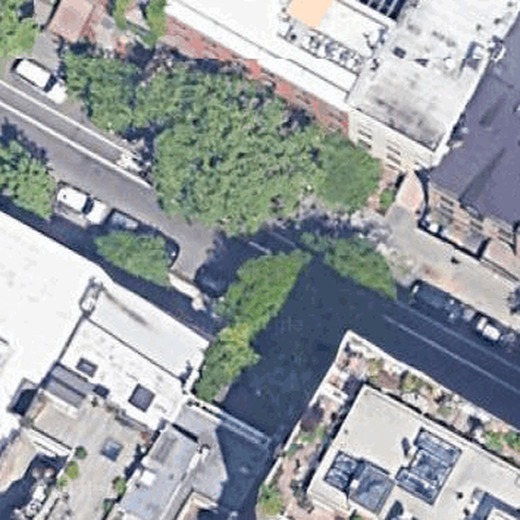}
\end{tikzpicture}%
}
\par\vspace{0.285cm}
\caption{\subcapaliasing}
\label{fig:challenges:aliasing}
\end{subfigure}
\caption{%
\subref{fig:challenges:degradation}~Existing fine-grained cross-view localization methods incur substantially larger errors as the aerial region expands from $35\times35$\,m to $400\times400$\,m.
\subref{fig:challenges:aliasing}~A large aerial region can contain multiple look-alike locations compatible with the same ground observation (inset). In this example, locations \textcolor{cLA}{$\bullet$}~A and \textcolor{cLB}{$\bullet$}~B share similar appearance with the true location (\textcolor{cGT}{$\star$}). The main differences lie in subtle spatial details, such as building outlines and the arrangement of surrounding vegetation.
}
\label{fig:challenges}
\end{figure}

To address these challenges, we propose \name{}, a search-and-verification framework for precise cross-view localization over large aerial regions. Rather than localizing directly over the full region, \name{} first performs a \emph{multi-hypothesis branch-and-zoom search} that explores the aerial region from coarse to fine. We use predictive uncertainty to expose distinct plausible hypotheses, retain them as separate search branches, and recursively zoom into their surroundings to access increasingly fine-grained evidence. By avoiding premature commitment under sparse cross-view correspondences, the search produces a compact shortlist of promising candidates.

Verification then focuses on resolving ambiguity among these look-alike candidates. To this end, we introduce the \emph{depth ring}, a view-shared geometric representation that encodes camera-centered distances to surrounding scene surfaces across horizontal azimuths. Its azimuth-wise depth pattern captures subtle variations in the surrounding spatial layout, forming a distinctive location-specific signature that can be compared across ground and aerial views. We derive the ground ring from estimated depth and train an aerial ring predictor to reconstruct the corresponding ring from each candidate-centered aerial patch.
A large vision-language model (LVLM) verifier then examines the overall ring patterns together with their visual context to select the final location.

To evaluate precise localization over large aerial regions, we construct VIGOR-Pro by automatically composing VIGOR aerial tiles into contiguous regions of approximately \(100\), \(200\), and \(400\)\,m. Across both same-area and cross-area evaluations, \name{} consistently outperforms representative end-to-end and retrieval-based localization methods. At \(400\)\,m, \name{} achieves mean localization errors of \(12.36\)\,m and \(19.81\)\,m on same-area and cross-area evaluation, respectively, compared with \(15.53\)\,m and \(26.38\)\,m for the strongest retrieval-based baseline.
Beyond this, further analysis shows that our depth ring can serve as explicit spatial evidence for improving zero-shot cross-view spatial reasoning in frontier multimodal models such as GPT-5.6 Terra and Gemini-3.7 Flash, highlighting its potential for broader applications.

In summary, our main contributions are as follows:
\begin{itemize}
    \item We study fine-grained cross-view localization beyond the conventional \emph{local-refinement} regime, extending the task to substantially larger aerial regions. We further construct VIGOR-Pro to enable systematic evaluation under this setting.

    \item We propose a \emph{multi-hypothesis branch-and-zoom search} that uses predictive uncertainty to preserve multiple plausible locations and progressively refine them, avoiding premature commitment under sparse cross-view correspondences.

    \item We introduce the \emph{depth ring}, a view-shared geometric representation that captures the fine-grained spatial structure surrounding a location. It enables effective disambiguation among look-alike locations and serves as explicit spatial evidence for cross-view reasoning.

\end{itemize}

\section{Problem Setup and Challenges}
\label{sec:setup}

\subsection{Problem Formulation}
\label{sec:problem_formulation}

Following the standard problem formulation of fine-grained cross-view localization~\citep{xia2022metric,xia2024ccvpe}, we are given a ground panorama $I_g$ and a geo-referenced aerial image $I_a$ depicting an aerial region that contains the camera location. In practice, $I_a$ can be obtained from an upstream localization system as a coarse geographic prior. Let $\Omega_a$ denote the image domain of $I_a$. Our goal is to estimate the camera position
\begin{equation}
    \hat{\mathbf{p}}=(\hat{x},\hat{y})\in\Omega_a.
\end{equation}


Unlike conventional fine-grained cross-view localization, we consider a broader aerial region to accommodate a coarser upstream localization prior. In this work, we consider the known-orientation setting, where the ground panorama has a known heading relative to the north-up aerial image, allowing us to isolate the challenges introduced by wide-area cross-view localization.

Since $I_a$ is geo-referenced, let $\mathcal{G}_a:\Omega_a\rightarrow\mathbb{R}^2$ map each image coordinate to its corresponding geographic coordinate. The localization error can then be measured directly in meters:
\begin{equation}
    e_{\mathrm{geo}}(\hat{\mathbf{p}},\mathbf{p}^{\star})
    =d_{\mathrm{geo}}\!\left(
        \mathcal{G}_a(\hat{\mathbf{p}}),
        \mathcal{G}_a(\mathbf{p}^{\star})
    \right),
\end{equation}
where $\mathbf{p}^{\star}\in\Omega_a$ denotes the ground-truth camera position and $d_{\mathrm{geo}}$ denotes the geographic distance.

\subsection{What Challenges Does Wide-Area Cross-View Localization Pose?}
\label{sec:challenges}

Expanding the aerial region introduces challenges beyond simply enlarging the coordinate range. We identify two that are particularly critical.

\paragraph{Sparse ground--aerial correspondences.}
A ground observation captures only the local surroundings of the camera. As the aerial extent grows, the mismatch in spatial extent between the ground observation and the aerial region becomes increasingly pronounced, leaving only a small portion of the region with evidence that corresponds to the ground view. As shown in Figure~\ref{fig:challenges:degradation}, expanding the aerial region can severely degrade existing end-to-end methods, particularly those relying on local cross-view alignment, including HC-Net~\citep{wang2023fine} and FG$^2$~\citep{xia2025fg2}.
This degradation stems from the fact that reliable cross-view evidence exists only within an unknown local area of the aerial region, making direct localization over the full region inherently challenging. The task therefore shifts from local refinement to a wide-area spatial search problem.

\paragraph{Visual ambiguity in ground--aerial matching.}
When cross-view localization is extended to larger aerial regions, recurring scene layouts and visual patterns can give rise to multiple look-alike locations. As shown in Figure~\ref{fig:challenges:aliasing}, such look-alike locations can share similar semantic content and visual layouts, making them difficult to distinguish based on appearance alone. Resolving this ambiguity requires fine-grained cross-view spatial reasoning: the ability to recognize and compare subtle spatial differences under large viewpoint disparities, including spatial configuration, relative geometry, and structural organization of scene elements. Recent studies show that multimodal models struggle with spatial reasoning, particularly egocentric--allocentric transformations~\citep{yang2025thinking}. These limitations become more pronounced across viewpoints, where models must establish spatial correspondence and preserve object and layout consistency~\citep{jeong2026xvr,liu2026cvsbench}.

\section{Methodology}
\label{sec:method}

\subsection{Overview of \name{}}

\begin{figure}[t]
    \centering
    \includegraphics[width=\linewidth, trim={0mm 0mm 3mm 5mm}, clip]{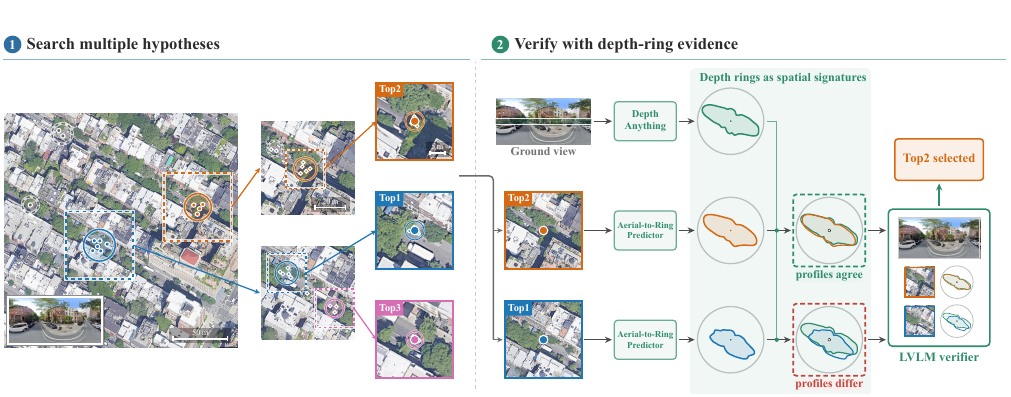}
    \caption{Overview of \name{}.
    \name{} first performs multi-hypothesis branch-and-zoom search over the aerial region, using predictive uncertainty to expose plausible hypotheses and recursively zoom into them to produce a compact candidate shortlist. It then resolves the remaining ambiguity with depth-ring-grounded verification. Ground observations and candidate aerial patches are mapped to depth rings that capture subtle details of the surrounding spatial layout. The correct candidate exhibits strong consistency with the ground observation, whereas a look-alike candidate does not. An LVLM verifier leverages this evidence to select the final location.}
    \label{fig:method_overview}
\end{figure}

Section~\ref{sec:challenges} leaves two coupled difficulties: cross-view correspondences over a large aerial region are sparse, and coarse visual evidence may point to multiple look-alike locations. Rather than aligning the ground view to the full aerial region in one step, we propose \name{}, a search-and-verification framework that first explores the aerial region for plausible locations and then disambiguates them using fine-grained spatial cues, as illustrated in Figure~\ref{fig:method_overview}.
\textit{Search} explores the aerial region from coarse to fine, retaining multiple plausible hypotheses and progressively zooming into their surroundings to produce a candidate shortlist. \textit{Verification} then distinguishes the remaining look-alike candidates. Since raw appearance is difficult to compare reliably across the ground--aerial viewpoint gap, we represent each view with a camera-centered \textit{depth ring}, a view-shared geometric representation that captures fine-grained spatial structure around a location. An LVLM verifier then leverages this evidence together with visual context to select the most consistent candidate.


\subsection{Multi-Hypothesis Branch-and-Zoom Search}
\label{sec:branch_zoom}

Sparse cross-view correspondences make a single prediction unreliable over a large aerial region. We therefore maintain multiple plausible hypotheses rather than committing prematurely. We represent the search as a tree, where each node corresponds to a location hypothesis and its associated search region. Starting from the root covering the full aerial region, we use predictive uncertainty to branch each node into multiple child hypotheses and zoom into their surroundings. Repeating this branch-and-zoom process progressively refines plausible locations while preserving alternative search paths. The resulting leaf hypotheses are ranked to produce a shortlist for verification.



\paragraph{Uncertainty-guided branching.}
Let node $v$ correspond to a search region $R_v\subseteq\Omega_a$, with $R_{\mathrm{root}}=\Omega_a$. We feed the aerial crop $I_a^{R_v}$ and $I_g$ to a localizer parameterized by $\theta$, which predicts a camera coordinate $\mathbf{u}\in R_v$. To expose multiple plausible modes, we draw $M$ stochastic predictions
\begin{equation}
    \mathcal{U}_v=\{\mathbf{u}_{v,m}\}_{m=1}^{M},
    \qquad
    \mathbf{u}_{v,m}\sim
    p_\theta(\mathbf{u}\mid I_g,I_a^{R_v}).
    \label{eq:uncertainty_sampling}
\end{equation}
We then apply single-linkage clustering~\citep{gower1969minimum} to group nearby predictions. Each retained cluster $C_{v'}\subseteq\mathcal{U}_v$ defines a child node $v'$, with its centroid as the corresponding hypothesis for a separate branch. Branching is therefore adaptive: concentrated predictions produce few branches, whereas multiple separated modes produce more. In this way, the search expands according to predictive uncertainty, allocating additional branches only when distinct hypotheses emerge.

\paragraph{Recursive zooming.}
For each retained hypothesis, we form a search region centered on it according to a predefined zoom schedule. Progressively narrowing the search region allows the localizer to access increasingly fine-grained evidence around each hypothesis. We repeat uncertainty-guided branching within each region, continuing this branch-and-zoom cycle until the final zoom level.


\paragraph{Hypothesis scoring and candidate shortlist.}
We score each hypothesis by aggregating the cluster sizes along its search path, based on the intuition that a promising hypothesis should remain well supported as the search progresses. For a hypothesis $\mathbf{h}$ at level $\ell$ with path $(v_0,v_1,\ldots,v_\ell)$, we define $S(\mathbf{h})=\sum_{t=1}^{\ell}|C_{v_t}|$, where $C_{v_t}$ is the cluster that produced node $v_t$. To bound the search cost, we rank hypotheses by $S$ after each level and retain at most $B$ for further expansion. At the final level, the top-$k$ leaf hypotheses $\{\mathbf{h}_i\}_{i=1}^{k}$ form the candidate shortlist for verification.

\subsection{Depth-Ring-Grounded Verification}
\label{sec:depth_ring_verify}

Search returns a shortlist of $k$ plausible locations, but residual ambiguity may remain among them. As noted in Section~\ref{sec:challenges}, distinguishing them requires fine-grained spatial reasoning across large viewpoint disparities. To resolve this ambiguity, we introduce the \emph{depth ring}, a view-shared geometric representation that encodes the estimated distance from a location to surrounding scene surfaces across horizontal azimuths. By expressing fine-grained spatial structure from both views in a shared representation, depth rings enable verification based on geometric consistency rather than direct appearance comparison across the viewpoint gap.


\paragraph{Depth ring as a spatial signature of a place.}
For a location $\mathbf{p}$, we define its depth ring as the distance from $\mathbf{p}$ to the first scene surface encountered along each horizontal direction:
\begin{equation}
    r_{\mathbf{p}}(\phi)
    =\inf\{\rho>0:\mathbf{p}+\rho\mathbf{d}(\phi)
    \text{ reaches a scene surface}\},
    \qquad \phi\in[0,2\pi),
    \label{eq:depth_ring}
\end{equation}
where $\phi$ denotes the azimuth, and
$\mathbf{d}(\phi)=(\cos\phi,\sin\phi)^{\top}$ is the corresponding unit vector in the horizontal plane. Sampling the ring at $A$ azimuths gives
$\mathbf{r}_{\mathbf{p}}
=[r_{\mathbf{p}}(\phi_1),\ldots,r_{\mathbf{p}}(\phi_A)]
\in\mathbb{R}^{A}$.

\begin{wrapfigure}{r}{0.45\linewidth}
    \vspace{-0.7\baselineskip}
    \centering
    \includegraphics[width=\linewidth, trim={0mm 4mm 1mm 1mm}, clip]{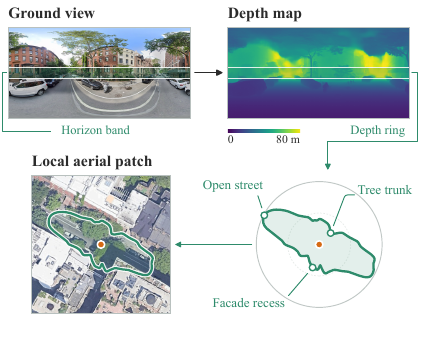}
    \caption{A depth ring summarizes azimuth-wise distances to surrounding scene surfaces in the horizontal plane. Its shape reveals subtle variations in the surrounding spatial layout, forming a compact spatial signature for cross-view comparison. The ring is derived from the ground observation and overlaid on the aerial patch only for visualization.}
    \label{fig:depthring}
    \vspace{-2.0\baselineskip}
\end{wrapfigure}



Although one-dimensional, the depth ring faithfully captures subtle variations in the surrounding spatial layout (see Figure~\ref{fig:depthring}). Its azimuth-wise depth pattern therefore forms a distinctive spatial signature of a location. As shown in Figure~\ref{fig:method_overview}, the aerial ring at the true location closely agrees with the ground ring, whereas a look-alike location with a different surrounding layout produces an inconsistent profile.
Since the depth ring is indexed by azimuth, cross-view comparison requires a common directional reference. Under the known-orientation setting, ground and aerial rings can be directly aligned; for unknown headings, the comparison can be extended by first estimating and compensating for the relative yaw.

This representation relates to classical isovist-based descriptions of surrounding visibility~\citep{benedikt1979isovist} and LiDAR place descriptors that compactly encode surrounding geometry~\citep{kim2018scancontext,lu2022ring}. However, since no range scans are directly available in our setting, depth rings must be inferred from ground and aerial observations.


\paragraph{Constructing rings from both views.}

\emph{Ground-to-ring.} The ground panorama observes the surrounding scene, with its horizontal coordinates naturally corresponding to azimuths. We use a pretrained metric depth model~\citep{yang2024depthanythingv2} to estimate a depth map $D_g$ from $I_g$. For each azimuth $\phi_a$, we then pool depth values within a narrow vertical band $\mathcal{B}(\phi_a)$ around the horizon:
\begin{equation}
    r_g(\phi_a)=
    \operatorname{Pool}\left\{
    D_g(x,y)\mid (x,y)\in\mathcal{B}(\phi_a)
    \right\},
    \qquad a=1,\ldots,A.
    \label{eq:ground_depth_ring}
\end{equation}

\emph{Aerial-to-ring.} 
Unlike the ground panorama, the aerial view does not directly provide azimuth-wise depth around a location. We therefore learn an aerial-to-ring predictor $f_\varphi$ that maps a local aerial patch $I_a^{\mathbf{p}}$ together with its specified location $\mathbf{p}$ to the corresponding azimuth-wise depth ring:
\begin{equation}
    \hat{\mathbf{r}}_a(\mathbf{p})
    =f_\varphi\left(I_a^{\mathbf{p}},\mathbf{p}\right).
    \label{eq:aerial_depth_ring}
\end{equation}
We implement $f_\varphi$ as azimuth-wise metric depth estimation. Training requires no additional annotation: for each ground--aerial pair with a known camera position, we construct the ground ring using Eq.~\eqref{eq:ground_depth_ring} and use it to supervise the aerial prediction at the corresponding location. Architectural and training details are provided in Appendix~\ref{app:ring_details}.

\paragraph{LVLM-based candidate verification.}
Although depth rings provide explicit spatial evidence, direct numerical comparison can be brittle under local depth errors. We therefore use a learned LVLM verifier to compare overall ring patterns rather than relying on point-wise agreement. For each candidate $\mathbf{h}_i$, we align the ground ring $\mathbf{r}_g$ and predicted aerial ring $\hat{\mathbf{r}}_a(\mathbf{h}_i)$ in a shared polar coordinate system, and provide them together with the ground panorama and the corresponding candidate aerial patch to the verifier. It then selects the most plausible candidate from $\{\mathbf{h}_i\}_{i=1}^{k}$ as the final prediction.

\subsection{Training Strategy for \name{}}
\label{sec:training}

\paragraph{Multi-scale training for search.}
We use a single cross-view localizer $p_\theta$ across all levels of branch-and-zoom, requiring it to operate over aerial regions at different scales. We therefore aggregate ground--aerial training pairs across scales and jointly train the localizer. Leveraging the visual understanding and flexible generation capabilities of LVLMs, we implement $p_\theta$ within an LVLM framework rather than using conventional cross-view localization models~\citep{xia2024ccvpe,xia2025fg2}. Given a ground panorama and an aerial image, the model is trained to generate the camera coordinate in text form using the ground-truth coordinate as supervision.


\paragraph{Hard-candidate training for verifier.}
After training the search model, we run multi-hypothesis search on the ground--aerial training pairs and use the retained hypotheses as hard candidates for verifier training. The verifier is fine-tuned to output the selected candidate in text form, using the candidate closest to the ground-truth location as supervision. Because these candidates are drawn from the hypotheses retained after search, they naturally preserve the residual visual ambiguity, exposing the verifier to challenging cases similar to those encountered at inference.

\section{Experiments}
\label{sec:experiments}

\subsection{Setup}
\label{sec:experimental_setup}

\paragraph{Dataset and evaluation protocol.}
Existing benchmarks for fine-grained cross-view localization are mainly designed for narrow geographic priors, where localization is performed within a relatively small aerial region. To evaluate precise localization over larger aerial regions, we construct VIGOR-Pro from VIGOR~\citep{zhu2021vigor} by composing neighboring aerial tiles into contiguous mosaics while preserving the original ground panoramas and camera locations. No additional imagery or manual annotation is introduced.
We construct three scales with approximate side lengths of $100$, $200$, and $400$\,m, corresponding to increasingly coarse geographic priors. Following the standard VIGOR protocol, we evaluate under both \emph{same-area} and \emph{cross-area} settings. The former uses all cities for both training and testing, while the latter trains on two cities and evaluates on the remaining two unseen cities. All experiments use the known-orientation setting, and we report mean and median localization errors in meters following prior work~\citep{xia2025fg2}. Detailed construction procedures and dataset statistics are provided in Appendix~\ref{app:data_protocol}.

\definecolor{ourscolor}{RGB}{235,243,250}
\begin{table}[t]
    \centering
    \caption{Main results on VIGOR-Pro under same-area and cross-area settings across increasingly large aerial regions. Mean and median localization errors are reported in meters. The best results are highlighted in bold, and the second-best results are underlined.}
    \label{tab:vigor_main}
    \resizebox{0.9\columnwidth}{!}{%
    \begin{tabular}{c l cc cc cc}
        \toprule
        \multirow{2}{*}{\textbf{Area}}
        & \multirow{2}{*}{\textbf{Method}}
        & \multicolumn{2}{c}{$\mathbf{100\,m}$}
        & \multicolumn{2}{c}{$\mathbf{200\,m}$}
        & \multicolumn{2}{c}{$\mathbf{400\,m}$} \\
        \cmidrule(lr){3-4}\cmidrule(lr){5-6}\cmidrule(lr){7-8}
        & & Mean $\downarrow$ & Median $\downarrow$
        & Mean $\downarrow$ & Median $\downarrow$
        & Mean $\downarrow$ & Median $\downarrow$ \\
        \midrule
        \multirow{12}{*}{\rotatebox[origin=c]{90}{Same-area}}
        & \multicolumn{7}{l}{\textit{End-to-End}} \\
        & HC-Net~\citep{wang2023fine}
            & 10.52 & 3.70 & 29.54 & 10.53 & 85.62 & 58.49 \\
        & CCVPE~\citep{xia2024ccvpe}
            & 9.63 & 2.11 & 14.07 & 2.64 & 27.90 & 3.48 \\
        & FG$^2$~\citep{xia2025fg2}
            & 9.76 & 4.63 & 18.38 & 8.14 & 38.41 & 17.00 \\
        & Qwen3-VL-4B
            & 5.36 & 2.27 & 12.16 & 3.52 & 23.03 & 3.93 \\
        \cmidrule(lr){2-8}
        & \multicolumn{7}{l}{\textit{Retrieval-Based}} \\
        & Sample4Geo~\citep{deuser2023sample4geo}
            & 23.84 & 20.85 & 22.80 & 16.85 & 23.84 & 15.89 \\
        & Sample4Geo+FG$^2$
            & 13.35 & 5.17 & 12.80 & 3.68 & \underline{15.53} & 4.72 \\
        & CIPER~\citep{jeon2026ciper}
            & 18.69 & 10.48 & 25.78 & 8.61 & 40.79 & 8.65 \\
        \cmidrule(lr){2-8}
        & \multicolumn{7}{>{\columncolor{ourscolor}}l}{\textit{Our Search-Based Framework}} \\
        & Single-Hypothesis
            & 5.00 & 2.00 & 10.29 & 2.27 & 21.42 & 2.46 \\
        & \textbf{\name{}} (Search-only)
            & \underline{4.32} & \textbf{1.65} & \underline{8.03} & \underline{1.82} & 15.60 & \underline{1.92} \\
        & \textbf{\name{}}
            & \textbf{4.02} & \textbf{1.65} & \textbf{6.57} & \textbf{1.79} & \textbf{12.36} & \textbf{1.88} \\
        \midrule
        \multirow{12}{*}{\rotatebox[origin=c]{90}{Cross-area}}
        & \multicolumn{7}{l}{\textit{End-to-End}} \\
        & HC-Net~\citep{wang2023fine}
            & 14.02 & 5.10 & 44.12 & 18.49 & 121.71 & 110.38 \\
        & CCVPE~\citep{xia2024ccvpe}
            & 19.29 & 3.58 & 41.65 & 4.67 & 94.18 & 7.69 \\
        & FG$^2$~\citep{xia2025fg2}
            & 10.77 & 5.22 & 21.68 & 10.23 & 62.21 & 35.56 \\
        & Qwen3-VL-4B 
            & 7.12 & 3.87 & 17.25 & 5.45 & 34.64 & 6.16 \\
        \cmidrule(lr){2-8}
        & \multicolumn{7}{l}{\textit{Retrieval-Based}} \\
        & Sample4Geo~\citep{deuser2023sample4geo}
            & 25.02 & 21.63 & 27.78 & 18.01 & 34.80 & 17.25 \\
        & Sample4Geo+FG$^2$
            & 14.00 & 4.86 & 17.66 & 3.36 & 26.38 & 4.78 \\
        & CIPER~\citep{jeon2026ciper}
            & 21.57 & 12.46 & 40.41 & 13.85 & 80.84 & 22.59 \\
        \cmidrule(lr){2-8}
        & \multicolumn{7}{>{\columncolor{ourscolor}}l}{\textit{Our Search-Based Framework}} \\
        & Single-Hypothesis
            & 5.79 & 2.48 & 12.90 & 2.29 & 29.19 & 2.40 \\
        & \textbf{\name{}} (Search-only)
            & \underline{4.95} & \underline{1.95} & \underline{11.01} & \underline{1.99} & \underline{23.99} & \underline{2.09} \\
        & \textbf{\name{}}
            & \textbf{4.81} & \textbf{1.94} & \textbf{9.31} & \textbf{1.96} & \textbf{19.81} & \textbf{2.04} \\
        \bottomrule
    \end{tabular}%
    }
\end{table}

\paragraph{Implementation details.}
The search model is initialized from Qwen3-VL-4B-Instruct~\citep{bai2025qwen3vl} and jointly trained on $50$, $100$, $200$, and $400$\,m regions to support localization across varying aerial extents. The additional $50$\,m regions are generated from the $100$\,m training regions and used only for multi-scale training.
During inference, we follow the $\{400,200,100,50\}$\,m zoom schedule starting from each evaluation scale. Unless specified otherwise, we draw $M=16$ predictions per node at temperature 1 and use a pruning budget of $B=4$.
After the final search stage, we retain the top-2 hypotheses for verification. Ground depth rings are constructed with Depth Anything V2~\citep{yang2024depthanythingv2}, and the aerial ring predictor is trained with ground-derived ring supervision.
The verifier is initialized from the search model and further fine-tuned on hard candidates generated under the same search configuration. Further implementation details are provided in Appendix~\ref{app:implementation}.

\paragraph{Baselines.}
We compare RingMatch against representative end-to-end and retrieval-based localization methods, and further report controlled variants within our search-based framework.
(1)~\emph{End-to-end localization} directly predicts the camera position from a ground--aerial pair. This category includes representative cross-view localization methods, such as HC-Net~\citep{wang2023fine}, CCVPE~\citep{xia2024ccvpe}, and FG$^2$~\citep{xia2025fg2}. We additionally report direct prediction results from our fine-tuned Qwen3-VL-4B search model.
(2)~\emph{Retrieval-based localization} treats the original aerial tiles composing each VIGOR-Pro mosaic as the retrieval gallery. We adopt Sample4Geo~\citep{deuser2023sample4geo}, a Sample4Geo+FG$^2$ cascade, and CIPER~\citep{jeon2026ciper}. Sample4Geo uses the center of the top-1 retrieved tile as the prediction; Sample4Geo+FG$^2$ further refines the location within that tile using FG$^2$, a recent strong method for local refinement, while CIPER jointly performs retrieval and local pose estimation.
(3)~\emph{Our search-based framework} progressively explores and zooms into the aerial region. Within this framework, we introduce Single-Hypothesis, an ablation variant that uses the same search model as \name{} but retains only one hypothesis per level.
Each existing end-to-end baseline is retrained independently at every VIGOR-Pro scale using its official configuration. For Sample4Geo and CIPER, we use the released checkpoints and follow their official inference procedures. Further details are provided in Appendix~\ref{app:baseline_details}.

\subsection{Main Results}
\label{sec:main_results}

\paragraph{Localization over increasingly large aerial regions.}
Table~\ref{tab:vigor_main} shows that conventional end-to-end localizers degrade sharply as the aerial region expands. On same-area evaluation, the mean error of CCVPE rises from $9.63$\,m at $100$\,m to $27.90$\,m at $400$\,m, with even larger increases for HC-Net and FG$^2$. In contrast, \name{} achieves the lowest errors across all scales, reaching $12.36/1.88$\,m mean/median error at $400$\,m, compared with $15.53/4.72$\,m for the strongest retrieval-based baseline.

\paragraph{Cross-area generalization.}
The advantage of \name{} persists when generalizing to unseen geographic areas. At $400$\,m, \name{} achieves mean/median errors of $19.81/2.04$\,m, outperforming the strongest retrieval-based baseline Sample4Geo+FG$^2$ with $26.38/4.78$\,m. This demonstrates the robustness of \name{} to geographic distribution shifts.

\paragraph{Broader regions demand multi-hypothesis exploration.}
Compared with Single-Hypothesis, \name{}~(Search-only) reduces the same-area mean error by $13.6\%$, $22.0\%$, and $27.2\%$ at $100$, $200$, and $400$\,m, respectively. The increasing gain shows that premature commitment becomes more costly as the aerial region expands, underscoring the need for multi-hypothesis exploration.

\paragraph{Depth-ring-grounded verification corrects costly mis-selections.}
Compared with search alone, verification further reduces the mean error at $400$\,m from $15.60$ to $12.36$\,m under same-area evaluation and from $23.99$ to $19.81$\,m under cross-area evaluation. In contrast, the corresponding median errors change only slightly. This suggests that verification primarily corrects costly mis-selections among plausible look-alike candidates rather than refining already accurate predictions.

\begin{table*}[t]
\centering
\caption{
Verifier ablation on VIGOR-Pro under same-area evaluation.
Mean localization error (m) is reported on all test samples (ALL) and the verification subset (VFY).
``Search init.'' denotes initialization from the trained search model.
}
\label{tab:verifier_ablation}
\small
\setlength{\tabcolsep}{4.5pt}
\renewcommand{\arraystretch}{1.08}

\scalebox{1.0}{%
\begin{tabular}{@{}l ccc cc cc cc@{}}
\toprule
& \multicolumn{3}{c}{Verifier components}
& \multicolumn{2}{c}{$100\,\mathrm{m}$}
& \multicolumn{2}{c}{$200\,\mathrm{m}$}
& \multicolumn{2}{c}{$400\,\mathrm{m}$} \\
\cmidrule(lr){2-4}
\cmidrule(lr){5-6}
\cmidrule(lr){7-8}
\cmidrule(lr){9-10}

& Local patch
& Depth ring
& Search init.
& ALL & VFY
& ALL & VFY
& ALL & VFY \\
\midrule

Search-only
& -- & -- & --
& 4.32 & 22.59
& 8.03 & 41.64
& 15.60 & 70.58 \\
\midrule

\multirow{3}{*}{w/ Verifier}
& $\checkmark$ & -- & --
& 4.60 & 26.89
& 8.67 & 48.11
& 17.05 & 81.06 \\

& $\checkmark$ & $\checkmark$ & --
& \underline{4.26} & \underline{21.67}
& \underline{7.36} & \underline{34.77}
& \underline{13.89} & \underline{58.17} \\

& $\checkmark$ & $\checkmark$ & $\checkmark$
& \textbf{4.02} & \textbf{17.59}
& \textbf{6.57} & \textbf{26.69}
& \textbf{12.36} & \textbf{47.12} \\
\bottomrule
\end{tabular}%
}
\end{table*}

\subsection{Analysis}

\begin{wraptable}{r}{0.46\linewidth}
    \centering
    \vspace{-0.23\baselineskip}
    \caption{Verification accuracy (\%) on VIGOR-Pro same-area verification samples. FT denotes task-specific fine-tuning. Ring-only selects the candidate with the higher Pearson correlation between its predicted aerial depth ring and the ground depth ring.}
    \label{tab:verification_accuracy}
    \small
    \setlength{\tabcolsep}{3.0pt}
    \renewcommand{\arraystretch}{1.04}
    \begin{tabular*}{\linewidth}{@{\extracolsep{\fill}}lccc@{}}
        \toprule
        Method & FT & $200\,\mathrm{m}$ & $400\,\mathrm{m}$ \\
        \midrule
        Random & -- & 50.00 & 50.00 \\
        Ring-only & -- & 67.29 & 69.42 \\
        \midrule
        GPT-5.6 Terra & -- & 56.93 & 59.47 \\
        \quad w/ depth ring & --
        & \textbf{67.58} {\scriptsize(+10.65)}
        & \textbf{66.21} {\scriptsize(+6.74)} \\
        Gemini-3.7 Flash & -- & 66.70 & 72.56 \\
        \quad w/ depth ring & --
        & \textbf{69.92} {\scriptsize(+3.22)}
        & \textbf{75.29} {\scriptsize(+2.73)} \\
        \midrule
        \name{} & $\checkmark$ & 81.92 & 82.81 \\
        \bottomrule
    \end{tabular*}
\end{wraptable}

\paragraph{Depth rings enable stronger zero-shot cross-view spatial reasoning.}
Table~\ref{tab:verification_accuracy} evaluates whether depth rings can help frontier multimodal models resolve visual ambiguity across ground--aerial views without task-specific fine-tuning. With visual evidence alone, GPT-5.6 Terra and Gemini-3.7 Flash remain limited in recognizing fine-grained spatial differences, with GPT-5.6 Terra performing only slightly above chance. Ring-only already achieves $67.29\%$ and $69.42\%$ accuracy at $200$ and $400$\,m, showing that depth rings carry discriminative spatial information. Providing depth-ring evidence improves GPT-5.6 Terra by $10.65$ and $6.74$ points and Gemini-3.7 Flash by $3.22$ and $2.73$ points. Notably, Gemini with depth-ring evidence further surpasses Ring-only, suggesting that depth rings serve not only as geometric descriptors for direct cross-view matching, but also as explicit spatial evidence for improving cross-view reasoning.

\paragraph{Depth-ring evidence improves candidate verification.}
Table~\ref{tab:verifier_ablation} isolates the verifier components. A verifier using visual patches alone fails to improve over search and can even degrade localization, confirming the difficulty of resolving look-alike candidates from appearance alone. Adding depth-ring evidence reduces the mean error on verification samples from $81.06$ to $58.17$\,m at $400$\,m, while initialization from the trained search model further lowers it to $47.12$\,m, showing the complementarity of explicit spatial evidence and cross-view representations learned during search.

\usetikzlibrary{pgfplots.groupplots}
\definecolor{paperOrange}{HTML}{D55E00}
\definecolor{paperGreen}{HTML}{009E73}
\definecolor{paperPink}{HTML}{D66EB4}
\definecolor{paperBlue}{HTML}{0072B2}
\pgfplotsset{searchaxis/.style={
 width=5.25cm,height=5.00cm,
 xmin=0.7,xmax=5.3,ymin=7,ymax=21,
 xtick={1,2,3,4,5},ytick={8,10,12,14,16,18,20},
 xlabel={Candidate count $k$},
 label style={font=\small},tick label style={font=\small},
 title style={yshift=-0.7ex,font=\small},
 grid=major,
 major grid style={line width=0.45pt,draw=gray!30},
 axis lines=box,
 legend style={font=\small,fill=white,fill opacity=0.92,text opacity=1,
 draw=gray!45,inner sep=2pt},legend pos=north east,
 legend cell align=left}}
\begin{figure}[t]
    \hspace*{-0.3cm}
    \centering
    \begin{tikzpicture}
\begin{groupplot}[searchaxis,group style={group size=3 by 1,horizontal sep=0.7cm,vertical sep=1.25cm,group name=searchplots}]
\nextgroupplot[title={(a) Level $L$},ylabel={Mean Distance Error (m) $\downarrow$},y label style={at={(axis description cs:-0.13,0.5)},font=\small,anchor=south}]
\addplot[color=paperOrange,mark=o,mark size=1.8pt,line width=1pt] coordinates {(1,19.488043496358) (2,12.490264000850) (3,11.383160847484) (4,11.383160847484) (5,11.383160847484)};
\addlegendentry{$L=1$}
\addplot[color=paperGreen,mark=o,mark size=1.8pt,line width=1pt] coordinates {(1,18.526827780674) (2,11.864444963633) (3,10.312054733103) (4,9.818233679060) (5,9.644192397463)};
\addlegendentry{$L=2$}
\addplot[color=paperPink,mark=o,mark size=1.8pt,line width=1pt] coordinates {(1,16.527771883374) (2,10.617791915291) (3,9.320825813837) (4,8.803637726689) (5,8.670006503135)};
\addlegendentry{$L=3$}
\addplot[color=paperBlue,mark=o,mark size=1.8pt,line width=1pt] coordinates {(1,15.60) (2,10.426106546195) (3,9.124446608028) (4,8.632364522232) (5,8.521661382657)};
\addlegendentry{$L=4$}
\nextgroupplot[title={(b) Samples $M$},yticklabels={}]
\addplot[color=paperOrange,mark=o,mark size=1.8pt,line width=1pt] coordinates {(1,16.881748198860) (2,12.943944137053) (3,12.188305442047) (4,11.904154295672) (5,11.827704707612)};
\addlegendentry{$M=8$}
\addplot[color=paperGreen,mark=o,mark size=1.8pt,line width=1pt] coordinates {(1,15.60) (2,10.426106546195) (3,9.124446608028) (4,8.632364522232) (5,8.521661382657)};
\addlegendentry{$M=16$}
\addplot[color=paperPink,mark=o,mark size=1.8pt,line width=1pt] coordinates {(1,15.545560407746) (2,9.765733873490) (3,8.367693861338) (4,7.945173221986) (5,7.840255849653)};
\addlegendentry{$M=24$}
\addplot[color=paperBlue,mark=o,mark size=1.8pt,line width=1pt] coordinates {(1,15.387209987713) (2,9.543130732992) (3,8.025855081192) (4,7.569422040370) (5,7.442849424495)};
\addlegendentry{$M=32$}
\nextgroupplot[title={(c) Budget $B$},yticklabels={}]
\addplot[color=paperOrange,mark=o,mark size=1.8pt,line width=1pt] coordinates {(1,16.041563471389) (2,11.361199401392) (3,11.059704542759) (4,10.916371718046) (5,10.906736501521)};
\addlegendentry{$B=2$}
\addplot[color=paperGreen,mark=o,mark size=1.8pt,line width=1pt] coordinates {(1,15.708941023092) (2,10.348947208346) (3,9.428837729437) (4,9.207351676484) (5,9.120443992991)};
\addlegendentry{$B=3$}
\addplot[color=paperPink,mark=o,mark size=1.8pt,line width=1pt] coordinates {(1,15.60) (2,10.426106546195) (3,9.124446608028) (4,8.632364522232) (5,8.521661382657)};
\addlegendentry{$B=4$}
\addplot[color=paperBlue,mark=o,mark size=1.8pt,line width=1pt] coordinates {(1,16.093198292324) (2,10.456816468357) (3,9.042126539025) (4,8.520012043208) (5,8.335077525865)};
\addlegendentry{$B=8$}
\end{groupplot}
    \end{tikzpicture}
\caption{Analysis of multi-hypothesis search on VIGOR-Pro under same-area evaluation at the $400$\,m scale. We report top-$k$ oracle mean localization error when varying (a) search level $L$, (b) sampling count $M$, and (c) pruning budget $B$, where the oracle selects the closest location among the top-$k$ candidates. Unless otherwise varied, we use $L=4$, $M=16$, and $B=4$.}

    \label{fig:search-ablation}
\end{figure}
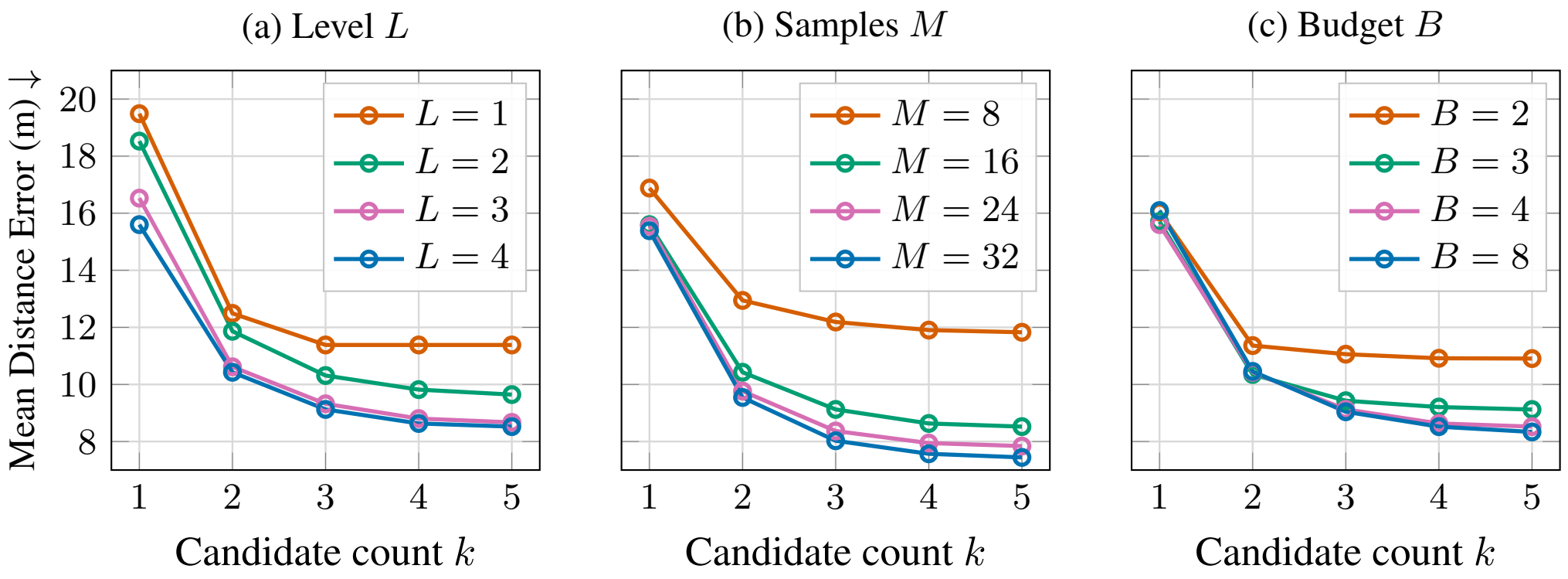

\paragraph{Multi-hypothesis search preserves promising locations.}
To isolate search quality from verification, Figure~\ref{fig:search-ablation} reports top-$k$ oracle mean error on VIGOR-Pro under same-area evaluation at $400$\,m. With the default setting ($L=4$, $M=16$, $B=4$), increasing $k$ from $1$ to $2$ reduces the error from $15.60$ to $10.43$\,m, while $k=5$ further reaches $8.52$\,m. The second candidate captures $73.0\%$ of the total gain from $k=1$ to $k=5$, supporting a compact top-2 shortlist for verification.

\begin{table}[t]
    \centering
    \caption{Generalization beyond VIGOR to four unseen cities at the $200$\,m scale.}
    \label{tab:beyond_vigor}
    \small
    \setlength{\tabcolsep}{3.2pt}
    \renewcommand{\arraystretch}{1.05}
    \resizebox{\columnwidth}{!}{%
    \begin{tabular}{l cc cc cc cc cc}
        \toprule
        \multirow{2}{*}{Method}
        & \multicolumn{2}{c}{London}
        & \multicolumn{2}{c}{Tokyo}
        & \multicolumn{2}{c}{S\~ao Paulo}
        & \multicolumn{2}{c}{Cape Town}
        & \multicolumn{2}{c}{Overall} \\
        \cmidrule(lr){2-3}
        \cmidrule(lr){4-5}
        \cmidrule(lr){6-7}
        \cmidrule(lr){8-9}
        \cmidrule(lr){10-11}
        & Mean $\downarrow$ & Median $\downarrow$
        & Mean $\downarrow$ & Median $\downarrow$
        & Mean $\downarrow$ & Median $\downarrow$
        & Mean $\downarrow$ & Median $\downarrow$
        & Mean $\downarrow$ & Median $\downarrow$ \\
        \midrule
        FG$^2$
        & 14.98 & 7.24
        & 36.26 & 20.87
        & 30.91 & 17.13
        & 23.52 & 12.41
        & 26.42 & 13.41 \\
        Sample4Geo+FG$^2$
        & \underline{9.47} & 2.22
        & 24.38 & 4.72
        & \underline{19.09} & 3.92
        & 14.14 & 2.82
        & 16.77 & 3.20 \\
        \midrule
        \textbf{\name{}} (Search-only)
        & 9.65 & \underline{1.23}
        & \underline{21.64} & \underline{1.04}
        & 19.95 & \underline{1.49}
        & \underline{9.57} & \underline{1.02}
        & \underline{15.20} & \underline{1.19} \\
        \textbf{\name{}}
        & \textbf{8.58} & \textbf{1.21}
        & \textbf{17.42} & \textbf{1.00}
        & \textbf{18.48} & \textbf{1.45}
        & \textbf{7.89} & \textbf{0.98}
        & \textbf{13.09} & \textbf{1.16} \\
        \bottomrule
    \end{tabular}%
    }
\end{table}

\paragraph{Sensitivity to search parameters.}
Search quality improves with deeper exploration and sufficient stochastic sampling. At $k=2$, increasing $L$ from $1$ to $4$ reduces the oracle error from $12.49$ to $10.43$\,m, while increasing $M$ from $8$ to $16$ reduces it from $12.94$ to $10.43$\,m, with diminishing gains at larger $M$. Performance is comparatively insensitive to the pruning budget when enough branches are retained, with $B=4$ and $8$ giving similar results. More results are provided in Appendix~\ref{app:extended_search}.

\paragraph{Generalization beyond VIGOR.}
To examine whether the observed gains extend beyond VIGOR, we further collect evaluation data from four cities outside the benchmark---London, Tokyo, S\~ao Paulo, and Cape Town---and evaluate at the $200$\,m scale. For all methods, we directly use models trained under the VIGOR same-area setting without target-city adaptation. As shown in Table~\ref{tab:beyond_vigor}, \name{} consistently outperforms both baselines across all four cities, achieving $13.09/1.16$\,m mean/median error overall compared with $16.77/3.20$\,m for Sample4Geo+FG$^2$. These results show that the gains of \name{} extend beyond the geographic and domain characteristics of VIGOR.
\newpage
\section{Related Work}
\label{sec:related_work}

\paragraph{Precise cross-view localization.}
Cross-view retrieval matches ground observations against geo-referenced aerial galleries~\citep{zhu2021vigor,zhu2022transgeo,deuser2023sample4geo}, while metric localization estimates continuous camera coordinates or poses within an aerial reference. Existing metric methods use dense pose predictions, geometric alignment, or fine-grained feature correspondences~\citep{wang2023fine,xia2024ccvpe,xia2025fg2}. Although effective for local refinement, these methods are typically evaluated with tightly bounded aerial priors. We study precise localization within substantially larger contiguous aerial regions, where sparse ground--aerial correspondences make direct localization increasingly difficult.

\paragraph{Wide-area cross-view localization and hierarchical search.}
UnifyGeo~\citep{song2025unifygeo} and CIPER~\citep{jeon2026ciper} integrate large-scale retrieval and metric localization within unified frameworks. Just Zoom In~\citep{erzurumlu2026zoom} and GeoMoE~\citep{fan2026manyscales} progressively narrow geographic search spaces through hierarchies of satellite cells or gallery entries. These approaches address related coarse-to-fine localization challenges, but emphasize retrieval--localization integration or hierarchical geographic narrowing. Our work focuses on recovering precise continuous coordinates within a given coarse aerial prior. We form spatially distinct hypotheses from stochastic coordinate predictions and recursively zoom into their neighborhoods, followed by explicit verification of the remaining ambiguous candidates.

\paragraph{Geometry and cross-view spatial reasoning.}
Cross-view matching has long leveraged geometric transformations, geometry-guided aggregation, and local correspondences~\citep{shi2019safa,lentsch2023slicematch,xia2025fg2}. Meanwhile, LVLM-based geolocation exploits scene semantics and geographic knowledge~\citep{li2024georeasoner,li2025globe}, yet fine-grained spatial reasoning remains challenging across viewpoints~\citep{jeong2026xvr,liu2026cvsbench}. To complement existing geometric matching approaches, we introduce the depth ring, a camera-centered representation of the surrounding spatial layout. By deriving ground rings from estimated depth and predicting corresponding aerial rings from overhead imagery, we provide explicit geometric evidence that enables an LVLM verifier to distinguish visually similar locations through fine-grained spatial differences. 

Appendix~\ref{app:related_work} provides an extended review of related work and a more detailed discussion of the connections between prior work and our approach.

\section{Conclusion}
We study fine-grained cross-view localization beyond the conventional local-refinement regime to better accommodate the coarse localization priors. To address the sparse cross-view correspondences and visual ambiguity arising over larger aerial regions, \name{} combines multi-hypothesis branch-and-zoom search with depth-ring-grounded verification, preserving plausible locations while exploiting fine-grained spatial structure for disambiguation. Experiments on VIGOR-Pro show consistent improvements across increasing aerial scales. Further analysis shows that depth rings provide explicit spatial evidence for cross-view reasoning in frontier multimodal models.

\bibliography{iclr2027_conference}
\bibliographystyle{iclr2027_conference}

\appendix
\clearpage
\section*{Appendix}
This appendix provides extended related work, dataset documentation, implementation and evaluation details, and additional qualitative and quantitative analyses for the main paper:
\begin{itemize}
    \item \S\,\ref{app:related_work} extends the related-work discussion on visual geo-localization, precise cross-view localization, geometry-based representations, and LVLM reasoning.
    \item \S\,\ref{app:data_protocol} documents VIGOR-Pro construction, coordinate transformations, benchmark scales, and dataset statistics, with representative wide-area examples in Figure~\ref{fig:vigor_pro_10x10_cases}.
    \item \S\,\ref{app:implementation} details training and inference for the search model, depth-ring predictor, and verifier, including supervision, optimization, and inference procedures.
    \item \S\,\ref{app:baseline_details} describes baseline adaptation and the common evaluation protocols for end-to-end, retrieval-based, and search-based methods.
    \item \S\,\ref{app:analysis_details} specifies the zero-shot cross-view reasoning experiments and the generalization evaluation beyond VIGOR, with verifier prompts in Figures~\ref{fig:verifier-prompt-ring} and~\ref{fig:verifier-prompt-no-ring}.
    \item \S\,\ref{app:depth_ring_qualitative} analyzes qualitative depth-ring examples and their ability to distinguish look-alike locations, illustrated in Figure~\ref{fig:depth_ring_cases}.
    \item \S\,\ref{app:extended_search} extends the analysis of search levels, sampling counts, and pruning budgets across all VIGOR-Pro scales and both evaluation settings in Figures~\ref{fig:search-level-full}--\ref{fig:search-budget-full}.
\end{itemize}

\section{Extended Related Work}
\label{app:related_work}

This section expands the discussion in Section~\ref{sec:related_work} with broader background and detailed comparisons.

\paragraph{Image geo-localization and cross-view geo-localization.}
Visual geo-localization spans a broad range of spatial extents. Global methods infer geographic locations through classification, retrieval, or joint image--coordinate representations~\citep{weyand2016planet,vo2017revisiting,vivanco2023geoclip}, while recent LVLM-based systems exploit semantic cues and geographic knowledge for reasoning-based image geo-localization~\citep{li2024georeasoner,jia2024g3,li2025globe}. When geo-referenced aerial imagery is available, cross-view geo-localization matches ground observations against an aerial gallery to identify the corresponding geographic location~\citep{hu2018cvmnet,shi2019safa,zhu2022transgeo,deuser2023sample4geo,ye2024epbev}. These methods provide broad geographic coverage and can serve as an upstream stage that narrows the localization prior.


\paragraph{Fine-grained cross-view localization.}
A complementary line of work focuses on precise metric localization within a local aerial reference. Existing methods predict continuous 2D positions or 3-DoF poses using dense pose distributions, geometric transformations, correspondence matching, or pose-conditioned features~\citep{xia2022metric,fervers2023uncertainty,lentsch2023slicematch,wang2023fine,xia2024ccvpe,xia2025fg2,park2026vird}. While these approaches achieve high metric accuracy, they largely operate in a local-refinement regime whose localization range is determined by the supplied aerial prior. On VIGOR~\citep{zhu2021vigor}, commonly used protocols assume only a small neighborhood around the camera~\citep{lentsch2023slicematch,xia2024ccvpe,xia2025fg2}. Our work instead studies precise cross-view localization when this upstream prior remains much broader.

\paragraph{Bridging global and fine-grained localization.}
Recent systems increasingly connect these two stages. UnifyGeo integrates retrieval, re-ranking, and metric localization, while CIPER jointly performs city-scale retrieval and local 3-DoF pose estimation~\citep{song2025unifygeo,jeon2026ciper}. Just Zoom In and GeoMoE instead perform hierarchical coarse-to-fine geographic narrowing~\citep{erzurumlu2026zoom,fan2026manyscales}. Their retrieval and narrowing stages are complementary to ours and can provide the coarse geographic prior assumed by our setting. Although hierarchical narrowing is related to our progressive search, these hierarchical methods primarily navigate predefined discrete cells or gallery entries. In contrast, we directly localize continuous camera coordinates within a contiguous aerial region, using predictive uncertainty to retain multiple plausible hypotheses and progressively refine them. Our goal is therefore to extend precise cross-view localization to broader geographic priors rather than replace the upstream localization stage.

\paragraph{Cross-view geometry and spatial representations.}
Across ground--aerial cross-view tasks, geometry has long been used to bridge the viewpoint gap through polar transformations, bird's-eye-view projections, homography estimation, geometry-guided matching, and point--plane correspondence~\citep{shi2019safa,shi2022beyond,lentsch2023slicematch,wang2023fine,ye2024epbev,xia2025fg2,park2026vird}. These designs support cross-view matching or pose estimation by making the two views more geometrically compatible. Our depth ring instead provides a view-shared geometric interface that brings spatial structure from both views into a common comparison space, exposing subtle layout differences among look-alike locations. It is conceptually related to vantage-point representations such as isovists and LiDAR place descriptors such as Scan Context and RING~\citep{benedikt1979isovist,kim2018scancontext,lu2022ring}. Unlike LiDAR descriptors, our image-only setting provides no direct range measurements. We therefore derive ground rings from estimated depth and learn to predict corresponding aerial rings from overhead imagery using ground-derived supervision.

\paragraph{LVLMs for geolocation and spatial reasoning.}
LVLMs have recently been explored for image geo-localization through semantic and geographic reasoning~\citep{li2024georeasoner,jia2024g3,li2025globe}. However, fine-grained spatial reasoning remains challenging for current multimodal models~\citep{wang2025spatial457,yang2025thinking}. Recent cross-view studies further highlight challenges in establishing correspondence and maintaining spatial consistency across viewpoints~\citep{jeong2026xvr,wang2026crossviewsuite,liu2026cvsbench}. Rather than relying on LVLMs to infer cross-view geometry directly from raw imagery, we use their predictive uncertainty for multi-hypothesis search and provide explicit depth-ring evidence for resolving look-alike locations.

\section{VIGOR-Pro Dataset Construction}
\label{app:data_protocol}

\begin{figure}[h]
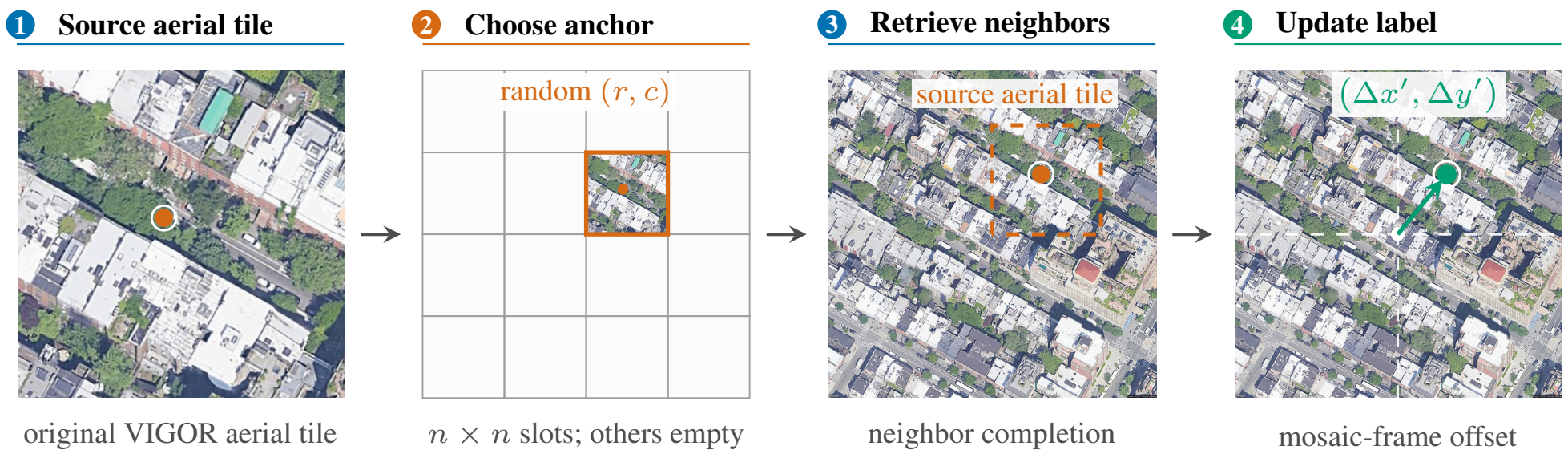

    \centering
    \providecolor{cFG2}{HTML}{0072B2}
    \providecolor{cCCVPE}{HTML}{009E73}
    \providecolor{cGT}{HTML}{D55E00}
    \begin{tikzpicture}[x=1cm,y=1cm,>=stealth,
        title/.style={font=\bfseries\scriptsize,anchor=west},
        note/.style={font=\scriptsize,align=center},
        flow/.style={->,line width=.8pt,black!70}]

        \def\panelW{2.70}
        \def\panelY{0.55}
        \def\panelH{2.70}

        \fill[cFG2] (0.03,3.63) circle (0.12);
        \node[font=\bfseries\tiny,text=white] at (0.03,3.63) {1};
        \node[title] at (0.22,3.63) {Source aerial tile};
        \draw[cFG2,line width=.5pt] (0,3.47) -- (2.70,3.47);
        \node[anchor=south west,inner sep=0] at (0,\panelY) {
            \includegraphics[width=\panelW cm,height=\panelH cm,
            trim={500bp 500bp 167bp 167bp},clip]
            {figure/preliminary/imgs_ny/aerial_region.jpg}};
        \fill[cGT] (1.215,2.035) circle (2.1pt);
        \draw[white,line width=.55pt] (1.215,2.035) circle (2.9pt);
        \node[note,text=black!70] at (1.35,0.24) {original VIGOR aerial tile};

        \draw[flow] (2.84,1.90) -- (3.17,1.90);

        \fill[cGT] (3.38,3.63) circle (0.12);
        \node[font=\bfseries\tiny,text=white] at (3.38,3.63) {2};
        \node[title] at (3.57,3.63) {Choose anchor};
        \draw[cGT,line width=.5pt] (3.35,3.47) -- (6.05,3.47);
        \fill[gray!3] (3.35,\panelY) rectangle ++(\panelW,\panelH);
        \foreach \i in {0,...,3}{
            \foreach \j in {0,...,3}{
                \draw[draw=black!38,line width=.35pt]
                    (3.35+0.675*\i,0.55+0.675*\j)
                    rectangle ++(0.675,0.675);
            }
        }
        \begin{scope}
            \clip (4.70,1.90) rectangle ++(0.675,0.675);
            \node[inner sep=0] at (5.0375,2.2375) {
                \includegraphics[width=.675cm,height=.675cm,
                trim={500bp 500bp 167bp 167bp},clip]
                {figure/preliminary/imgs_ny/aerial_region.jpg}};
        \end{scope}
        \draw[cGT,line width=1.0pt]
            (4.70,1.90) rectangle ++(0.675,0.675);
        \fill[cGT] (5.004,2.271) circle (1.25pt);
        \node[note,text=cGT,fill=white,fill opacity=.92,
            text opacity=1,inner sep=1pt] at (4.70,3.07) {random $(r,c)$};
        \node[note,text=black!70] at (4.70,0.24) {$n\times n$ slots; others empty};

        \draw[flow] (6.19,1.90) -- (6.52,1.90);

        \fill[cFG2] (6.73,3.63) circle (0.12);
        \node[font=\bfseries\tiny,text=white] at (6.73,3.63) {3};
        \node[title] at (6.92,3.63) {Retrieve neighbors};
        \draw[cFG2,line width=.5pt] (6.70,3.47) -- (9.40,3.47);
        \node[anchor=south west,inner sep=0] at (6.70,\panelY) {
            \includegraphics[width=\panelW cm,height=\panelH cm]
            {figure/preliminary/imgs_ny/aerial_region.jpg}};
        \draw[cGT,dashed,line width=.95pt]
            (8.05,1.90) rectangle ++(0.90,0.90);
        \fill[cGT] (8.455,2.394) circle (2.1pt);
        \draw[white,line width=.55pt] (8.455,2.394) circle (2.9pt);
        \node[note,text=cGT,fill=white,fill opacity=.90,
            text opacity=1,inner sep=1pt] at (8.24,3.06) {source aerial tile};
        \node[note,text=black!70] at (8.05,0.24) {neighbor completion};

        \draw[flow] (9.54,1.90) -- (9.87,1.90);

        \fill[cCCVPE] (10.08,3.63) circle (0.12);
        \node[font=\bfseries\tiny,text=white] at (10.08,3.63) {4};
        \node[title] at (10.27,3.63) {Update label};
        \draw[cCCVPE,line width=.5pt] (10.05,3.47) -- (12.75,3.47);
        \node[anchor=south west,inner sep=0] at (10.05,\panelY) {
            \includegraphics[width=\panelW cm,height=\panelH cm]
            {figure/preliminary/imgs_ny/aerial_region.jpg}};
        \draw[white,dashed,line width=.6pt,opacity=.9]
            (11.40,\panelY) -- (11.40,3.25);
        \draw[white,dashed,line width=.6pt,opacity=.9]
            (10.05,1.90) -- (12.75,1.90);
        \fill[white] (11.40,1.90) circle (1.6pt);
        \fill[cCCVPE] (11.805,2.394) circle (2.3pt);
        \draw[white,line width=.6pt] (11.805,2.394) circle (3.1pt);
        \draw[->,cCCVPE,line width=1pt]
            (11.40,1.90) -- (11.77,2.35);
        \node[note,text=cCCVPE,fill=white,fill opacity=.9,
            text opacity=1,inner sep=1pt] at (11.57,3.05)
            {$\left(\Delta x',\Delta y'\right)$};
        \node[note,text=black!70] at (11.40,0.24)
            {mosaic-frame offset};
    \end{tikzpicture}
    \caption{VIGOR-Pro construction retains the original camera location while
    enlarging its aerial context. The source tile is assigned to a sampled
    lattice cell, geographic neighbors complete the overlapping mosaic, and
    the camera offset is transformed into the mosaic frame. The empty
    grid in Step~2 denotes logical tile slots; completed tiles overlap by half
    their width.}
    \label{fig:vigor_pro_construction}
\end{figure}

VIGOR-Pro turns an existing VIGOR panorama--aerial pair into a large-area
localization example. The construction changes only the aerial context and coordinate frame, while preserving the original panorama and camera location. It introduces neither newly collected imagery nor manual
location labels. Figure~\ref{fig:vigor_pro_construction} illustrates the overall construction process.

\subsection{Construction Protocol}
\label{app:vigor_construction}

\paragraph{Tile lattice and mosaic geometry.}
VIGOR aerial tiles are $640\times640$ pixels with $50\%$ overlap, hence a
320-pixel lattice stride. An $n\times n$ mosaic has native side length
\begin{equation}
    L_n = 640 + 320(n-1).
    \label{eq:vigor_mosaic_side}
\end{equation}
Tiles are pasted in a deterministic order at this stride. In overlap regions,
later tiles overwrite earlier pixels; no interpolation or blending is applied.
Thus every output pixel is copied from an original VIGOR tile.

\paragraph{Geographic retrieval and rejection.}
Each VIGOR query is associated with one positive and three semi-positive aerial
tiles. We independently use each of these four tiles as a source anchor. For a
query--source pair, the source tile is placed at a uniformly sampled lattice
cell $(r,c)\in\{0,\ldots,n-1\}^2$. Expected centers for the remaining cells are
obtained by stepping on the city-specific VIGOR lattice. A spatial index returns
the nearest tile, which is accepted only when the geodesic distance between its
center and the expected center is below $10\,$m. A mosaic is valid only if all
$n^2$ cells pass this test. The query is retained when at least one of its four
source anchors yields a complete mosaic.

\paragraph{Coordinate transformation.}
Let $(\Delta x,\Delta y)$ denote the camera offset from the native aerial tile center, and let $q=(n-1)/2$. With image $x$ increasing eastward and image $y$
increasing southward, the offset in the native mosaic frame is
\begin{equation}
    \Delta x' = \Delta x+320(q-c),
    \qquad
    \Delta y' = \Delta y+320(r-q).
    \label{eq:vigor_pro_offset}
\end{equation}
We express the target in an image-size-independent $[0,1000]^2$ coordinate
system as
\begin{equation}
    x = 1000\left(\frac{1}{2}-\frac{\Delta x'}{L_n}\right),
    \qquad
    y = 1000\left(\frac{1}{2}+\frac{\Delta y'}{L_n}\right).
    \label{eq:vigor_pro_normalized_coord}
\end{equation}
The inverse used by evaluation is
\begin{equation}
    \Delta x'=L_n\left(\frac{1}{2}-\frac{x}{1000}\right),
    \qquad
    \Delta y'=L_n\left(\frac{y}{1000}-\frac{1}{2}\right).
    \label{eq:vigor_pro_inverse_coord}
\end{equation}
Equations~\eqref{eq:vigor_pro_normalized_coord} and
\eqref{eq:vigor_pro_inverse_coord} form an exact round trip before decimal
rounding. The center of a mosaic maps to $(500,500)$. Algorithm~\ref{alg:vigor-pro} summarizes the complete construction procedure,
including geographic retrieval, mosaic validation, stitching, and coordinate
transformation.

\begin{paperalgorithm}{Construct one query's VIGOR-Pro examples}
\label{alg:vigor-pro}
\begin{tabbing}
\hspace{1.2em}\=\hspace{1.2em}\=\kill
\textbf{Input:} panorama $I_g$; source tuples
$\{(T_j,\Delta x_j,\Delta y_j)\}_{j=1}^{4}$; grid size $n$.\\
\textbf{Output:} zero or more complete mosaic--label pairs.\\[2pt]
\textbf{for} source anchor $j=1,\ldots,4$ \textbf{do}\\
\> Sample anchor cell $(r_j,c_j)$ uniformly.\\
\> For every cell, compute its expected geographic center and retrieve the
nearest city tile.\\
\> Reject anchor $j$ if any center discrepancy is at least $10\,$m.\\
\> Otherwise stitch at stride 320 and transform the label with
Eqs.~\eqref{eq:vigor_pro_offset}--\eqref{eq:vigor_pro_normalized_coord}.\\
\textbf{end for}\\
Return all valid pairs; discard $I_g$ only if none remain.
\end{tabbing}
\end{paperalgorithm}

\subsection{Benchmark Scales}
\label{app:vigor_configurations}

\begin{figure}[p]
    \centering
    \includegraphics[
        height=0.85\textheight,
        trim={220bp 0bp 220bp 0bp},
        clip
    ]{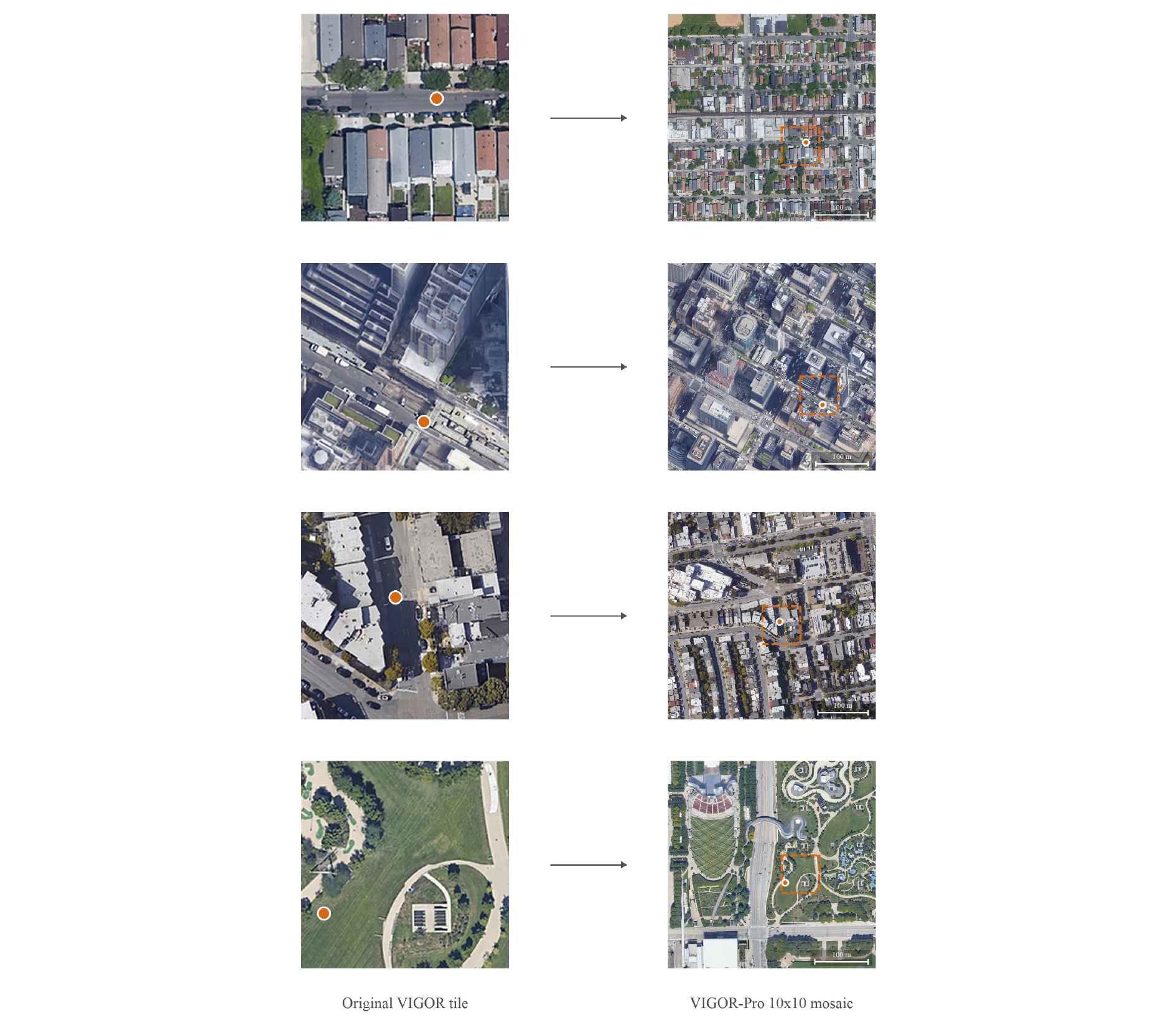}
\caption{Examples of VIGOR-Pro at the $400$\,m scale.
Each row compares an original VIGOR aerial tile (left) with the corresponding
VIGOR-Pro $10\times10$ mosaic (right). The orange point denotes the ground-camera
location, and the dashed orange box indicates the footprint of the original tile
within the enlarged aerial region.}
\label{fig:vigor_pro_10x10_cases}
\end{figure}

We instantiate $n\in\{2,5,10\}$, yielding native mosaic resolutions
$L_n\in\{960,1920,3520\}$ pixels. Using the city-specific ground sampling
distance (GSD) reported by SliceMatch~\citep{lentsch2023slicematch}, the mean
side lengths are approximately $106\,$m, $213\,$m, and $390\,$m, which we
refer to as the $100$, $200$, and $400$\,m scales, respectively. The
city-specific GSD values and corresponding physical side lengths are
reported in Table~\ref{tab:vigor_ground_resolution}, and representative examples are shown in Figure~\ref{fig:vigor_pro_10x10_cases}.

\paragraph{Dataset split.}
The benchmark inherits the official VIGOR query split. Neighbor retrieval is
performed only within the query city and does not alter the original
training/test city partition. All mosaics originating from the same panorama
remain in the same split.

\paragraph{Dataset statistics.}
Table~\ref{tab:vigor_scale_summary} reports statistics after complete-mosaic
filtering. \emph{Original panoramas} counts ground panoramas in the source
VIGOR annotations, and \emph{original pairs} counts their associated
panorama--tile pairs before mosaic construction. Because each panorama
has four source tiles, the 105,214 original panoramas yield 420,856 original
pairs at every scale. \emph{Complete mosaics} counts original pairs for which
all $n^2$ aerial tiles are recovered. The completion rate is the number of
complete mosaics divided by the number of original pairs.
\emph{Retained panoramas} counts unique ground panoramas with at least one
complete mosaic.

\begin{table}[h]
    \centering
    \caption{Aerial-region side lengths induced by the source GSD and native mosaic geometry. Values are side lengths in meters.}
    \label{tab:vigor_ground_resolution}
    \small
    \setlength{\tabcolsep}{4.5pt}
    \begin{tabular}{lrrrr}
        \toprule
        City & GSD (m/pixel) & $100$\,m & $200$\,m & $400$\,m \\
        \midrule
        New York      & 0.113248 & 108.72 & 217.44 & 398.63 \\
        Seattle       & 0.100817 &  96.78 & 193.57 & 354.88 \\
        San Francisco & 0.118141 & 113.42 & 226.83 & 415.86 \\
        Chicago       & 0.111262 & 106.81 & 213.62 & 391.64 \\
        \midrule
        Mean          &          & 106.43 & 212.87 & 390.25 \\
        \bottomrule
    \end{tabular}
\end{table}

\begin{table}[h]
    \centering
    \caption{VIGOR-Pro geometry and construction statistics after
    complete-mosaic filtering.}
    \label{tab:vigor_scale_summary}
    \small
    \setlength{\tabcolsep}{3.5pt}
    \renewcommand{\arraystretch}{1.08}
    \resizebox{\linewidth}{!}{%
    \begin{tabular}{lrrrrrrrr}
        \toprule
        Scale & Grid & Native & Mean side & Original & Original & Complete & Completion & Retained \\
        & $n\times n$ & resolution & (m) & panoramas & pairs & mosaics & rate (\%) & panoramas \\
        \midrule
        $100$\,m & $2\times2$   & $960^2$  & 106 & 105,214 & 420,856 & 417,375 & 99.17 & 105,214 \\
        $200$\,m & $5\times5$   & $1920^2$ & 213 & 105,214 & 420,856 & 400,740 & 95.22 & 104,098 \\
        $400$\,m & $10\times10$ & $3520^2$ & 390 & 105,214 & 420,856 & 372,532 & 88.52 & 101,053 \\
        \bottomrule
    \end{tabular}%
    }
\end{table}

\section{Implementation Details}
\label{app:implementation}

This section describes the training and inference details of the multi-scale search model, depth-ring
predictor, and learned verifier, including their supervision, inputs, and optimization settings.

\subsection{Search Model: Training and Inference}
\label{app:search_details}

\paragraph{Supervision format.}
Each supervised example contains exactly two images in a fixed order---the ground panorama followed by the aerial region---and one user/assistant turn. The template below gives both roles; the two image inputs are inserted in this order.

\begin{figure}[h]
\centering
\setlength{\fboxsep}{9pt}
\fbox{%
\begin{minipage}{\dimexpr\linewidth-2\fboxsep-2\fboxrule\relax}
\small\raggedright
\textbf{User}\par\smallskip
\texttt{<image><image>}\par
\texttt{Given a ground-level panoramic image and a satellite image, please
predict the coordinates [x, y] of the ground-level camera location in the
satellite image.}
\par\medskip
\textbf{Assistant}\par\smallskip
\texttt{[x, y]}
\end{minipage}}
\end{figure}

The assistant target is the bare string \texttt{[$x$, $y$]}, with each value rounded to one decimal place in the normalized coordinate system of Eq.~\eqref{eq:vigor_pro_normalized_coord}; no explanation or auxiliary token is appended. Keeping the same coordinate grammar at every scale permits one model to operate throughout the tree.

\paragraph{Multi-scale training set.}
Separate search models are trained for same-area and cross-area evaluation using their respective training splits. The training mixture contains direct VIGOR-Pro examples with approximately $100$, $200$, and $400\,$m coverage. We additionally introduce a $50\,$m scale to provide finer-grained supervision for the search stage. These examples are derived by cropping $1/2$-scale regions from the $100\,$m training examples.
To simulate realistic search refinement, the crop center is displaced from the ground-truth location using independent zero-mean Gaussian jitter with a standard deviation equal to $0.1$ of the crop half-size. The jitter is truncated to ensure that the ground-truth location remains inside the crop whenever the parent label lies within the image. At an image boundary, the crop is translated inward rather than reduced in size. Let $(a_x,a_y,b_x,b_y)$ denote the crop box in the normalized parent-image coordinates, and $(x,y)$ the ground-truth camera location. The in-crop label is $x'=1000(x-a_x)/(b_x-a_x)$ and $y'=1000(y-a_y)/(b_y-a_y)$.


\paragraph{Optimization.}
We initialize from Qwen3-VL-4B-Instruct. During training, only the language model parameters are updated, while the vision tower and multimodal projector remain frozen. All models are trained on 8$\times$ NVIDIA A100 (80GB) GPUs. Table~\ref{tab:search_training_config} summarizes the optimization configuration.

\begin{table}[t]
    \centering
    \caption{Search-model training configuration.}
    \label{tab:search_training_config}
    \setlength{\tabcolsep}{5.5pt}
    \begin{tabular}{ll@{\qquad}ll}
        \toprule
        Configuration & Value & Configuration & Value \\
        \midrule
        Initialization & Qwen3-VL-4B-Instruct
            & Optimizer & AdamW \\
        Trainable & Language model
            & Batch size & 256 \\
        Frozen & Vision, projector
            & Learning rate & $1\times10^{-5}$ \\
        Training scales & $50/100/200/400$\,m
            & Model input & $480^2/512^2/512^2/768^2$\,px \\
        Training epochs & 2
            & LR schedule & Cosine \\
        Warmup ratio & 0.1
            & & \\
        \bottomrule
    \end{tabular}
\end{table}

\begin{table}[t]
    \centering
    \caption{Branch-and-zoom inference schedule. The search-region schedule
    denotes the approximate side length of the aerial region presented at
    each level.}
    \label{tab:search_inference_config}
    \small
    \setlength{\tabcolsep}{5pt}
    \begin{tabular}{ccc}
        \toprule
        Initial scale & Search-region schedule (m) & Model-input size (px) \\
        \midrule
        $100$\,m
            & $100\!\to\!50$
            & $512\!\to\!480$ \\
        $200$\,m
            & $200\!\to\!100\!\to\!50$
            & $512\!\to\!512\!\to\!480$ \\
        $400$\,m
            & $400\!\to\!200\!\to\!100\!\to\!50$
            & $768\!\to\!512\!\to\!512\!\to\!480$ \\
        \bottomrule
    \end{tabular}
\end{table}

\paragraph{Branch-and-zoom inference.}
Table~\ref{tab:search_inference_config} summarizes the branch-and-zoom schedule used at each evaluation scale.
At each node, we draw $M$ textual coordinate samples. The coordinates
are clustered by single linkage in the normalized crop-local coordinate
system, using a Euclidean distance threshold of 50. Clusters with fewer than
two members are discarded, and at most the three largest clusters are
retained. Each retained cluster is represented by its centroid, which defines
a child hypothesis. After expanding all active nodes at each level, we rank
the resulting hypotheses using the path-aggregated cluster-size score defined
in Section~\ref{sec:branch_zoom} and retain at most $B$ for further
expansion. For each retained hypothesis, we form the corresponding child search region and extract its aerial crop directly from the original mosaic. If the crop
would exceed an image boundary, the search region is translated inward
without changing its size. To express predictions from different search
regions in a common coordinate frame, we record each region as
$(a_x,a_y,b_x,b_y)$ in the normalized root-mosaic coordinate system.
A local prediction $\mathbf{u}=(u_x,u_y)$ is then mapped back to the root
frame as
\begin{equation}
  \mathcal{T}(\mathbf{u})=
  \left(
  a_x+\frac{u_x}{1000}(b_x-a_x),
  a_y+\frac{u_y}{1000}(b_y-a_y)
  \right).
\end{equation}
After the final level, the leaf hypotheses are ranked by the same score, and
the top-$k$ are retained for verification.

\subsection{Depth Ring: Construction, Learning, and Inference}
\label{app:ring_details}

\begin{figure}[t]
\centering
\includegraphics[width=\linewidth]{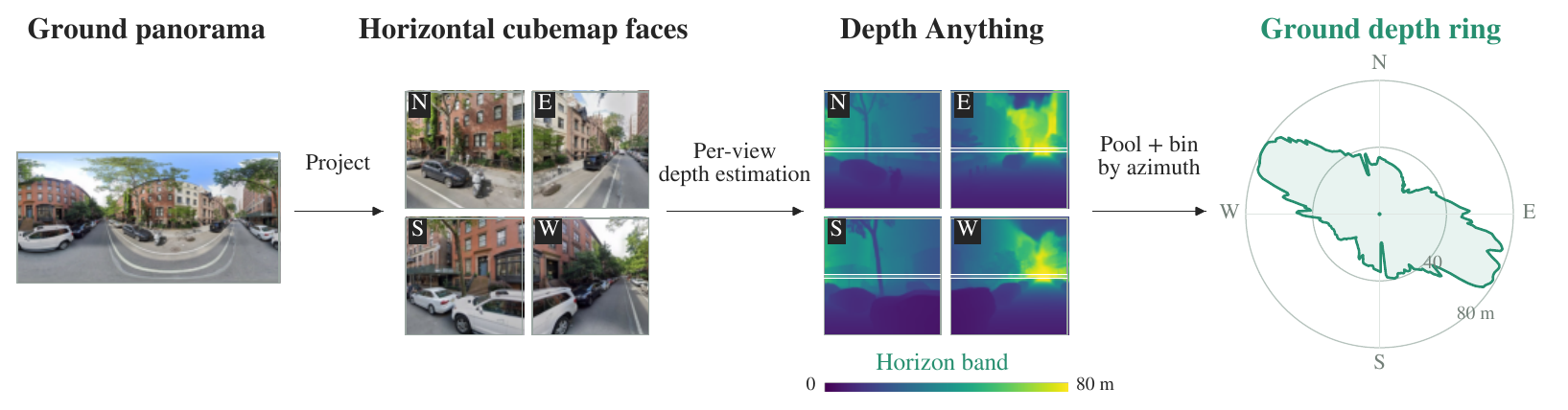}
\caption{Ground depth-ring construction.
The panorama is projected into four rectilinear views using
equirectangular-to-perspective (E2P) projection. Depth Anything V2 Metric Outdoor Large estimates depth for each view, from which we pool the
highlighted horizon band and map the resulting samples to their world azimuths.
Samples from all views are then aggregated into the $A$ uniformly spaced
azimuths of the ground depth ring.}
\label{fig:ground_ring_cubemap}
\end{figure}

\paragraph{Ground panorama to depth ring.}
VIGOR panoramas are oriented with the center facing north. We sample the depth ring at $A$ uniformly spaced clockwise azimuths $\phi_a=360^\circ(a-1)/A$, starting from the leftmost panorama direction as $\phi=0^\circ$. Since Depth Anything V2 is designed for perspective images, we first apply equirectangular-to-perspective (E2P) projection to render four $512\times512$ rectilinear views with view directions $\psi_j\in\{0^\circ,90^\circ,180^\circ,270^\circ\}$, pitch $0^\circ$, and horizontal field of view $90^\circ$.
We then apply Depth Anything V2 Metric Outdoor Large to each view independently. Eq.~\eqref{eq:ground_depth_ring} is implemented by mapping each perspective-view column back to its panorama azimuth. For a view of width $W_f$ and focal length $f=W_f/[2\tan(45^\circ)]$, column $x$ of view $j$ corresponds to
\begin{equation}
 \phi_j(x)=
 \left[\psi_j+\tan^{-1}\!\left(\frac{x-W_f/2}{f}\right)\right]
 \bmod 360^\circ .
 \label{eq:cubemap_azimuth}
\end{equation}
For each target azimuth $\phi_a$, we pool the estimated metric depth values in the corresponding horizon band to obtain $r_g(\phi_a)$ as in
Eq.~\eqref{eq:ground_depth_ring}.
For robustness to isolated depth artifacts, each perspective-view column is
represented by the median estimated depth within a $\pm8$-row band around the horizontal center. After mapping the samples from all four views to panorama azimuths using Eq.~\eqref{eq:cubemap_azimuth}, we aggregate them into the $A$ uniformly spaced azimuths of the depth ring, averaging samples assigned to the same azimuth. The Metric Outdoor Large model uses an $80\,$m depth range, preserving a common nominal metric scale across panoramas. Figure~\ref{fig:ground_ring_cubemap} illustrates the conversion from the panorama to the four perspective views and the resulting depth ring construction.

\begin{figure}[t]
\centering
\includegraphics[width=\linewidth]{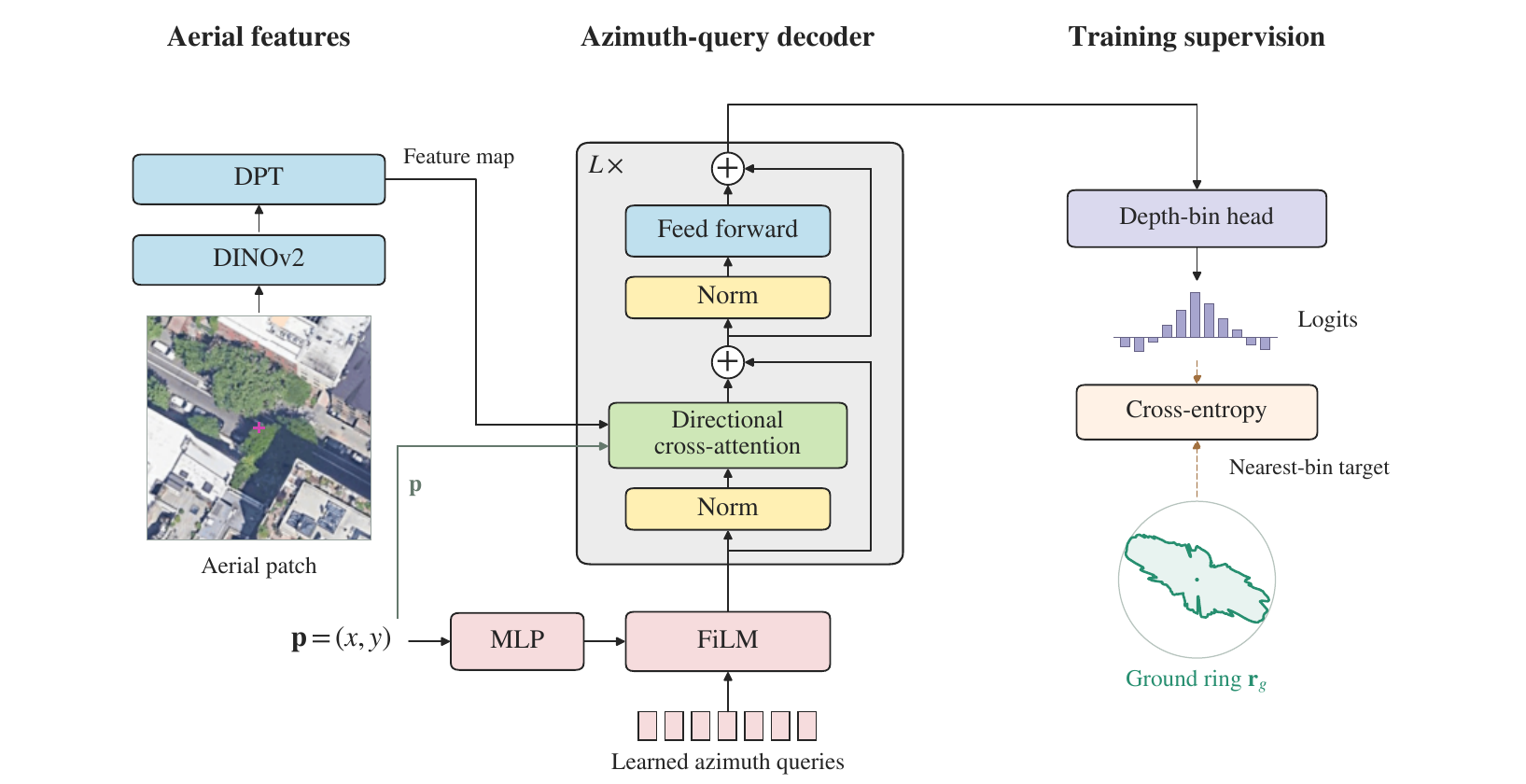}
\caption{Architecture of the aerial-to-ring predictor.
DINOv2 and DPT encode the local aerial patch into a dense feature map.
The candidate location $\mathbf{p}$ conditions the $A$ azimuth queries through
FiLM modulation. For each azimuth, a directional beam is sampled from
$\mathbf{p}$, and the corresponding query attends only to features along that
beam. Each updated query predicts a metric-depth class, producing the aerial
depth ring over all azimuths. During training, the predictor is supervised by
the ground depth ring extracted from the corresponding panorama.}
\label{fig:aerial_ring_predictor}
\end{figure}

\begin{table}[t]
    \centering
    \caption{Aerial-to-ring model and training configuration.}
    \label{tab:ring_config}
    \setlength{\tabcolsep}{5pt}
    \begin{tabular}{ll@{\quad}ll}
        \toprule
        Configuration & Value & Configuration & Value \\
        \midrule
        \multicolumn{4}{l}{\textbf{Model}} \\
        Aerial patch & $640^2$ &  FFN width & 1,024 \\
        Azimuth samples $A$ & 360 & Depth classes $K$ & 80 \\
        Metric depth range & $[0,80]\,$m & Backbone & frozen DINOv2-B \\
        DPT output width & 256 & Cross-attn. layers / heads & 4 / 8 \\
        \midrule
        \multicolumn{4}{l}{\textbf{Training}} \\
        Training loss & azimuth-wise CE & Optimizer & AdamW \\
        Batch size & 64 & Learning rate & $1\times10^{-4}$ \\
        Weight decay & 0.05 & Warmup & 500 steps \\
        LR schedule & cosine & Training epochs & $1$ \\
        \bottomrule
    \end{tabular}
\end{table}

\paragraph{Training recipe of the aerial-to-ring predictor.}
Figure~\ref{fig:aerial_ring_predictor} illustrates the architecture of the aerial-to-ring predictor. It uses a frozen DINOv2-Base encoder~\citep{oquab2023dinov2} with multi-level features from transformer blocks 3, 6, 9, and 12, which are fused by a DPT decoder~\citep{ranftl2021vision} into a dense aerial feature map. It maintains $A$ learned azimuth queries. The candidate coordinate $\mathbf{p}$, represented in the normalized local-patch coordinate system, is encoded by an MLP and used to condition these queries through feature-wise linear modulation (FiLM)~\citep{perez2018film}, while its mapped feature-map location serves as the origin for directional sampling. For each azimuth $\phi_a$, we sample a thin directional beam from the aerial feature map starting at $\mathbf{p}$ to exclude irrelevant context outside that direction. Each azimuth query is then updated by four cross-attention blocks using only the features sampled along its corresponding beam. The prediction head maps each query to logits $z_{a,k}$ over $K$ metric-depth classes corresponding to uniformly spaced depth values in $[0,80]$\,m. Each ground-ring value is quantized to its nearest metric-depth class and used as the cross-entropy target. The predictor is supervised by the ground depth ring constructed above and trained with azimuth-wise cross-entropy.

Training uses the VIGOR-Pro training sets at the $100$, $200$, and $400$\,m scales, with separate predictors trained for same-area and cross-area evaluation using their respective training splits. Each panorama--aerial pair provides one example. From each aerial mosaic, we extract a $640\times640$ local patch around the annotated camera location. At image boundaries, the crop is shifted inward while preserving its size, and the camera coordinate is updated accordingly.

At inference, the same aerial preprocessing is applied to each search hypothesis. For each azimuth, we select the highest-scoring depth class and use its corresponding depth value as the prediction.

Table~\ref{tab:ring_config} summarizes the training and model configuration of
the aerial-to-ring predictor.

\subsection{Verifier: Hard-Candidate Training and Inference}
\label{app:verifier_details}

\paragraph{Hard-candidate training set.}
Separate verifiers are trained for same-area and cross-area evaluation using their respective training splits.
Hard candidates from the $100$, $200$, and $400$\,m scales are combined for verifier training. For each ground--aerial pair in the training split, we run the same
branch-and-zoom procedure used at test time, with $M=16$ stochastic predictions
per node, sampling temperature $1$, pruning budget $B=4$, and the corresponding
zoom schedule. We then take the two highest-scoring hypotheses at the final
search level.
We discard examples for which fewer than two valid hypotheses are available or
the two candidates are separated by less than $10$\,m. The latter removes
near-duplicate alternatives that provide little useful selection supervision.
For each retained pair, the two hypotheses are randomly assigned to Candidate A
and Candidate B, preventing the
verifier from exploiting candidate order. The candidate with the
smaller Euclidean distance to the ground-truth camera position is used as the
supervision target. Importantly, the competing candidate is generated by the search model itself rather than independently sampled, exposing the verifier to plausible hard alternatives that remain after search.

\paragraph{Multimodal input and supervision.}
Each verifier example contains six images in a fixed order:
(1) the ground panorama;
(2) the full aerial region with Candidates A and B marked;
(3--4) local aerial patches around Candidates A and B; and
(5--6) candidate-specific depth-ring comparisons.
For each candidate, the ground depth ring and the predicted aerial ring are
rendered together in a common polar representation. The two candidate-specific
comparisons use the same radial scale, enabling direct comparison of their
geometric consistency. The candidate coordinates in the root aerial region are
additionally provided in the normalized coordinate system.
The supervision target contains only
\texttt{\textbackslash boxed\{A\}} or
\texttt{\textbackslash boxed\{B\}}, without an auxiliary rationale or
coordinate-regression target. We provide the complete prompt in Figure~\ref{fig:verifier-prompt-ring}.

\paragraph{Optimization and inference.}
We initialize the verifier from the trained search model, retaining the cross-view representations learned during multi-scale search training. As in search-model training, the vision tower and multimodal projector are frozen, and only the language-model parameters are updated. All models are trained on 8$\times$ NVIDIA A100 (80GB) GPUs. Table~\ref{tab:verifier_config} summarizes the optimization configuration.

At inference, branch-and-zoom search first returns the two highest-scoring final hypotheses. We apply verification only when the two candidates are at least $10$\,m apart; otherwise, the top-ranked search hypothesis is retained as the final prediction. For verified samples, we construct the corresponding verifier input. The verifier performs greedy decoding
with temperature $0$ and selects Candidate A or B. The selected hypothesis,
expressed in the root aerial coordinate frame, is returned directly as the
final localization result.

\begin{table}[h]
    \centering
    \caption{Verifier training configuration.}
    \label{tab:verifier_config}
    \setlength{\tabcolsep}{4.5pt}
    \begin{tabular}{ll@{\qquad}ll}
        \toprule
        Configuration & Value & Configuration & Value \\
        \midrule
        Initialization & Search model
            & Training epochs & 2 \\
        Trainable & Language model
            & Optimizer & AdamW \\
        Frozen & Vision, projector
            & Learning rate & $1\times10^{-5}$ \\
        Batch size & 256
            & LR schedule & Cosine \\
        Warmup ratio & 0.1
            & Candidate shortlist & 2 \\
        Candidate crop & $640\times640$
            &  &  \\
        \bottomrule
    \end{tabular}
\end{table}
\section{Baseline Adaptation and Evaluation Protocols}
\label{app:baseline_details}

\paragraph{Common evaluation protocol.}
For each VIGOR-Pro scale and split, all methods are evaluated on the same retained examples and receive the same underlying geographic prior defined by the corresponding aerial mosaic. End-to-end and search-based methods operate on the complete contiguous mosaic, whereas retrieval-based methods search over the original VIGOR tiles composing the same region. All predictions are mapped to the native mosaic coordinate frame before evaluation and converted to metric errors using the city-specific GSD in Table~\ref{tab:vigor_ground_resolution}.

\paragraph{End-to-end localization.}
HC-Net~\citep{wang2023fine}, CCVPE~\citep{xia2024ccvpe}, and
FG$^2$~\citep{xia2025fg2} are retrained independently at the $100$, $200$,
and $400$\,m scales of VIGOR-Pro. At each scale, the complete aerial mosaic is provided as input and resized according to the preprocessing of the corresponding official implementation. Camera labels and model predictions are transformed consistently between the resized input and the native mosaic coordinate frame. Apart from these scale-dependent adaptations, we follow the official architectures, objectives, training recipes, and inference settings. For Qwen3-VL-4B, we directly evaluate our fine-tuned search model by using its single-pass prediction without applying the search procedure.

\paragraph{Retrieval-based localization.}
For each VIGOR-Pro example, the retrieval gallery consists only of the original VIGOR tiles composing the corresponding mosaic, rather than the full VIGOR gallery. This preserves the native tile scale used by the retrieval models while restricting the search to the same geographic region available to the other localization paradigms. For Sample4Geo~\citep{deuser2023sample4geo}, we use the officially released checkpoint and inference configuration, taking the center of the top-ranked tile as the localization prediction. For Sample4Geo+FG$^2$, the same retrieved tile is further processed by an FG$^2$ model fine-tuned on the original VIGOR dataset for within-tile localization, and the resulting prediction is mapped to the native mosaic coordinate frame according to the tile position. For CIPER~\citep{jeon2026ciper}, we likewise use the officially released checkpoint and inference configuration on the same gallery, and map the pose prediction associated with the selected aerial tile to the native mosaic coordinate frame.

\paragraph{Our search-based framework.}
Single-Hypothesis uses the same trained search model and inference configuration as \name{}, but retains only the highest-scoring hypothesis at each search level for subsequent expansion. We also report \name{}~(Search-only), which follows the same search procedure but directly selects the top-ranked hypotheses without applying verification.
\section{Experimental Details for Additional Analyses}
\label{app:analysis_details}

\subsection{Zero-Shot Cross-View Spatial Reasoning} \label{app:zero_shot_reasoning} 

We evaluate zero-shot cross-view spatial reasoning on the VIGOR-Pro same-area verification samples at the $200$ and $400$\,m scales. Following the verification protocol in Appendix~\ref{app:verifier_details}, the two highest-scoring final search hypotheses form Candidates A and B, and samples with candidate separation below $10$\,m are excluded. The candidate closer to the ground-truth camera location is treated as the correct answer. We evaluate GPT-5.6 Terra and Gemini-3.7 Flash without task-specific fine-tuning. For each model, we compare two input settings on the same candidate pairs. In the visual-only setting, the model receives the ground panorama, the full aerial region with both candidates marked, and a local aerial patch centered on each candidate, using the prompt in Figure~\ref{fig:verifier-prompt-no-ring}. The depth-ring setting additionally provides the ground ring paired with the predicted aerial ring for each candidate, together with the corresponding instructions in Figure~\ref{fig:verifier-prompt-ring}. We report candidate selection accuracy. As a direct geometric baseline, Ring-only selects the candidate whose predicted aerial ring has the higher Pearson correlation with the ground ring. We also report the verifier in \name{}, which is fine-tuned for this task, on the same samples for reference.

\subsection{Generalization Beyond VIGOR}
\label{app:beyond_vigor_setup}

\paragraph{Dataset construction.}
We construct an additional evaluation set of $4{,}000$ panorama--aerial pairs from four cities outside the VIGOR benchmark: London, Tokyo, S\~ao Paulo, and Cape Town, with $1{,}000$ examples per city. Ground panoramas are collected from Google Street View, sampled at least $75$\,m apart, and converted to equirectangular images with the central viewing direction pointing north. Aerial imagery is collected through the Google Maps Static API. Following the same $5\times5$ mosaic geometry used at the $200$\,m VIGOR-Pro scale, each aerial region is composed of 25 overlapping $640\times640$ satellite images queried at zoom level 20 with $50\%$ overlap, yielding a $1920\times1920$ mosaic. To avoid center bias, the camera-containing tile is evenly distributed across the 25 grid positions, with 40 examples per position in each city. No examples from these cities are used for task-specific training.

\begin{wraptable}{r}{0.60\linewidth}
    \centering
    \vspace{-1.0\baselineskip}
    \caption{Aerial-region side lengths induced by the city-specific GSD and the native mosaic geometry.}
    \label{tab:beyond_vigor_ground_resolution}
    \small
    \setlength{\tabcolsep}{10pt}
    \renewcommand{\arraystretch}{1.05}
    \begin{tabular}{lrr}
        \toprule
        City & GSD (m/pixel) & Side length (m) \\
        \midrule
        London       & 0.092883 & 178.33 \\
        Tokyo        & 0.121224 & 232.75 \\
        S\~ao Paulo  & 0.136834 & 262.72 \\
        Cape Town    & 0.123851 & 237.79 \\
        \midrule
        Mean         &          & 227.90 \\
        \bottomrule
    \end{tabular}
    \vspace{-1.0\baselineskip}
\end{wraptable}

\paragraph{Evaluation protocol.}
We evaluate FG$^2$, Sample4Geo+FG$^2$ and \name{} without target-city fine-tuning or adaptation, using models trained under the VIGOR same-area setting. Localization errors are measured in the native mosaic coordinate system and converted to meters using the corresponding city-specific ground sampling distances, which are summarized in Table~\ref{tab:beyond_vigor_ground_resolution}.

\section{Qualitative Analysis of Depth Rings}
\label{app:depth_ring_qualitative}

Figure~\ref{fig:depth_ring_cases} provides qualitative examples of depth rings
for cross-view matching. Across diverse scene layouts, aerial rings predicted
near the ground-truth camera location preserve the characteristic spatial
structure of the corresponding ground ring, including its dominant directions,
relative extents, and local shape variations. In contrast, rings predicted at
incorrect locations exhibit clear structural discrepancies, even when the
candidate aerial regions contain similar roads or surrounding buildings.
These examples illustrate how the depth ring maps ground and aerial observations
into a shared representation of surrounding spatial layout, providing
discriminative geometric evidence beyond direct cross-view appearance matching.
\section{Extended Analysis of Multi-Hypothesis Search}
\label{app:extended_search}

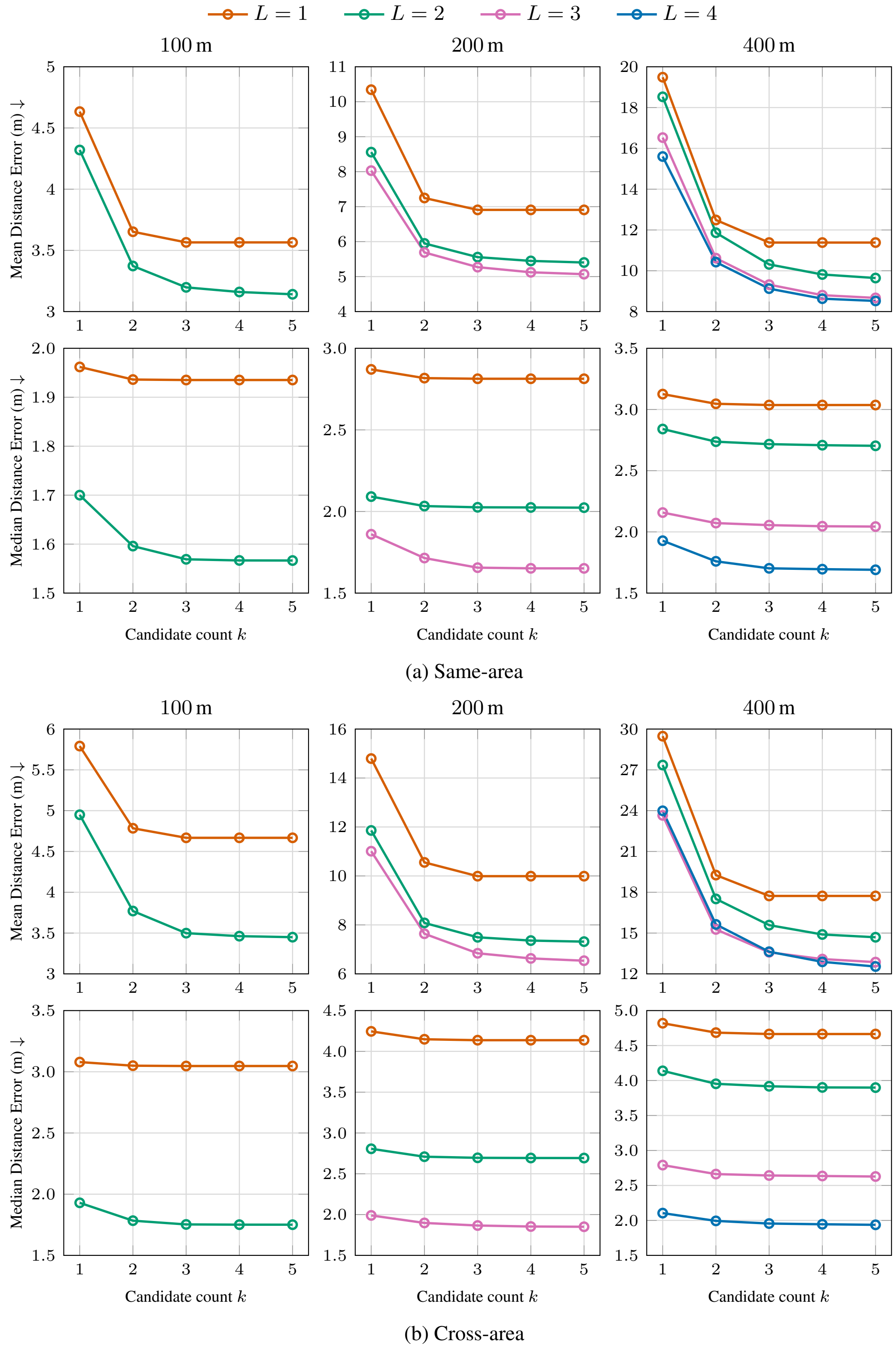
\begin{figure}[!ht]
\centering
\pgfplotslegendfromname{searchlevellegend}
\vspace{0.08cm}
\begin{subfigure}[t]{\linewidth}
\centering
\hspace*{-0.6cm}
\begin{tikzpicture}
\begin{groupplot}[searchaxis,width=5.25cm,height=5.25cm,label style={font=\scriptsize},tick label style={font=\scriptsize},group style={group size=3 by 2,horizontal sep=0.7cm,vertical sep=0.55cm}]
\nextgroupplot[ymin=3,ymax=5,ytick={3,3.5,4,4.5,5},title={$100$\,m},xlabel={},ylabel={Mean Distance Error (m) $\downarrow$},y label style={at={(axis description cs:-0.125,0.5)},font=\scriptsize,anchor=south}]
\addplot[color=paperOrange,mark=o,mark size=1.8pt,line width=1pt] coordinates {(1,4.633584738401) (2,3.652233619568) (3,3.565315827949) (4,3.565315827949) (5,3.565315827949)};
\addplot[color=paperGreen,mark=o,mark size=1.8pt,line width=1pt] coordinates {(1,4.32) (2,3.372402373300) (3,3.197918496108) (4,3.159989079638) (5,3.141394495653)};
\nextgroupplot[ymin=4,ymax=11,ytick={4,5,6,7,8,9,10,11},title={$200$\,m},xlabel={}]
\addplot[color=paperOrange,mark=o,mark size=1.8pt,line width=1pt] coordinates {(1,10.344490281124) (2,7.247990217025) (3,6.907913918685) (4,6.907913918685) (5,6.907913918685)};
\addplot[color=paperGreen,mark=o,mark size=1.8pt,line width=1pt] coordinates {(1,8.558594832295) (2,5.949272758023) (3,5.556793486656) (4,5.448764483844) (5,5.403483181404)};
\addplot[color=paperPink,mark=o,mark size=1.8pt,line width=1pt] coordinates {(1,8.03) (2,5.691253085257) (3,5.270550943366) (4,5.123494151369) (5,5.069070661640)};
\nextgroupplot[ymin=8,ymax=20,ytick={8,10,12,14,16,18,20},title={$400$\,m},xlabel={},legend to name=searchlevellegend,legend columns=-1,legend style={font=\small,fill=none,draw=none,inner sep=0pt,legend columns=-1,/tikz/every even column/.append style={column sep=0.45cm}}]
\addplot[color=paperOrange,mark=o,mark size=1.8pt,line width=1pt] coordinates {(1,19.488043496358) (2,12.490264000850) (3,11.383160847484) (4,11.383160847484) (5,11.383160847484)};
\addlegendentry{$L=1$}
\addplot[color=paperGreen,mark=o,mark size=1.8pt,line width=1pt] coordinates {(1,18.526827780674) (2,11.864444963633) (3,10.312054733103) (4,9.818233679060) (5,9.644192397463)};
\addlegendentry{$L=2$}
\addplot[color=paperPink,mark=o,mark size=1.8pt,line width=1pt] coordinates {(1,16.527771883374) (2,10.617791915291) (3,9.320825813837) (4,8.803637726689) (5,8.670006503135)};
\addlegendentry{$L=3$}
\addplot[color=paperBlue,mark=o,mark size=1.8pt,line width=1pt] coordinates {(1,15.60) (2,10.426106546195) (3,9.124446608028) (4,8.632364522232) (5,8.521661382657)};
\addlegendentry{$L=4$}
\nextgroupplot[ymin=1.5,ymax=2.0,ytick={1.5,1.6,1.7,1.8,1.9,2.0},scaled y ticks=false,yticklabel style={font=\scriptsize,/pgf/number format/fixed,/pgf/number format/fixed zerofill,/pgf/number format/precision=1},ylabel={Median Distance Error (m) $\downarrow$},y label style={at={(axis description cs:-0.125,0.5)},font=\scriptsize,anchor=south}]
\addplot[color=paperOrange,mark=o,mark size=1.8pt,line width=1pt] coordinates {(1,1.961900540286) (2,1.936143598208) (3,1.935204823552) (4,1.935204823552) (5,1.935204823552)};
\addplot[color=paperGreen,mark=o,mark size=1.8pt,line width=1pt] coordinates {(1,1.700046448810) (2,1.595970154412) (3,1.569001726999) (4,1.566659371036) (5,1.566494691972)};
\nextgroupplot[ymin=1.5,ymax=3.0,ytick={1.5,2.0,2.5,3.0},scaled y ticks=false,yticklabel style={font=\scriptsize,/pgf/number format/fixed,/pgf/number format/fixed zerofill,/pgf/number format/precision=1}]
\addplot[color=paperOrange,mark=o,mark size=1.8pt,line width=1pt] coordinates {(1,2.870726106199) (2,2.817189215977) (3,2.813275778910) (4,2.813275778910) (5,2.813275778910)};
\addplot[color=paperGreen,mark=o,mark size=1.8pt,line width=1pt] coordinates {(1,2.091221636386) (2,2.033259948835) (3,2.025698031739) (4,2.024734956990) (5,2.023444024987)};
\addplot[color=paperPink,mark=o,mark size=1.8pt,line width=1pt] coordinates {(1,1.860633237955) (2,1.714787141381) (3,1.655824351188) (4,1.651750108028) (5,1.651500397897)};
\nextgroupplot[ymin=1.5,ymax=3.5,ytick={1.5,2,2.5,3,3.5},scaled y ticks=false,yticklabel style={font=\scriptsize,/pgf/number format/fixed,/pgf/number format/fixed zerofill,/pgf/number format/precision=1}]
\addplot[color=paperOrange,mark=o,mark size=1.8pt,line width=1pt] coordinates {(1,3.125746292452) (2,3.046566405311) (3,3.036407795484) (4,3.036407795484) (5,3.036407795484)};
\addplot[color=paperGreen,mark=o,mark size=1.8pt,line width=1pt] coordinates {(1,2.840350440829) (2,2.736929919529) (3,2.716861164979) (4,2.708289382671) (5,2.703189235327)};
\addplot[color=paperPink,mark=o,mark size=1.8pt,line width=1pt] coordinates {(1,2.157979146546) (2,2.071945438267) (3,2.055292762094) (4,2.046260807459) (5,2.043338319060)};
\addplot[color=paperBlue,mark=o,mark size=1.8pt,line width=1pt] coordinates {(1,1.927912952708) (2,1.759645780394) (3,1.702350313192) (4,1.695391442547) (5,1.690386071844)};
\end{groupplot}
\end{tikzpicture}
\caption{Same-area}
\label{fig:search-level-samearea}
\end{subfigure}
\vspace{0.25cm}
\begin{subfigure}[t]{\linewidth}
\centering
\hspace*{-0.6cm}
\begin{tikzpicture}
\begin{groupplot}[searchaxis,width=5.25cm,height=5.25cm,label style={font=\scriptsize},tick label style={font=\scriptsize},group style={group size=3 by 2,horizontal sep=0.7cm,vertical sep=0.55cm}]
\nextgroupplot[ymin=3,ymax=6,ytick={3,3.5,4,4.5,5,5.5,6},title={$100$\,m},xlabel={},ylabel={Mean Distance Error (m) $\downarrow$},y label style={at={(axis description cs:-0.125,0.5)},font=\scriptsize,anchor=south}]
\addplot[color=paperOrange,mark=o,mark size=1.8pt,line width=1pt] coordinates {(1,5.791332141678) (2,4.784251804256) (3,4.667221264410) (4,4.667221264410) (5,4.667221264410)};
\addplot[color=paperGreen,mark=o,mark size=1.8pt,line width=1pt] coordinates {(1,4.95) (2,3.769998105402) (3,3.498888877936) (4,3.462195412388) (5,3.449691138383)};
\nextgroupplot[ymin=6,ymax=16,ytick={6,8,10,12,14,16},title={$200$\,m},xlabel={}]
\addplot[color=paperOrange,mark=o,mark size=1.8pt,line width=1pt] coordinates {(1,14.794325517640) (2,10.553609002347) (3,9.990629675919) (4,9.990629675919) (5,9.990629675919)};
\addplot[color=paperGreen,mark=o,mark size=1.8pt,line width=1pt] coordinates {(1,11.858034101898) (2,8.081644923994) (3,7.494190407310) (4,7.359872262253) (5,7.316558654229)};
\addplot[color=paperPink,mark=o,mark size=1.8pt,line width=1pt] coordinates {(1,11.01) (2,7.638748511999) (3,6.840987895277) (4,6.631419270730) (5,6.537599874119)};
\nextgroupplot[ymin=12,ymax=30,ytick={12,15,18,21,24,27,30},title={$400$\,m},xlabel={}]
\addplot[color=paperOrange,mark=o,mark size=1.8pt,line width=1pt] coordinates {(1,29.475445003424) (2,19.263522832619) (3,17.729857471739) (4,17.729857471739) (5,17.729857471739)};
\addplot[color=paperGreen,mark=o,mark size=1.8pt,line width=1pt] coordinates {(1,27.353729630759) (2,17.511675536773) (3,15.584248088016) (4,14.896756285444) (5,14.692376830353)};
\addplot[color=paperPink,mark=o,mark size=1.8pt,line width=1pt] coordinates {(1,23.653037230873) (2,15.269036218395) (3,13.586330376058) (4,13.087666518382) (5,12.864972883034)};
\addplot[color=paperBlue,mark=o,mark size=1.8pt,line width=1pt] coordinates {(1,23.99) (2,15.637527531406) (3,13.628519284830) (4,12.883404732705) (5,12.539866095711)};
\nextgroupplot[ymin=1.5,ymax=3.5,ytick={1.5,2.0,2.5,3.0,3.5},scaled y ticks=false,yticklabel style={font=\scriptsize,/pgf/number format/fixed,/pgf/number format/fixed zerofill,/pgf/number format/precision=1},ylabel={Median Distance Error (m) $\downarrow$},y label style={at={(axis description cs:-0.125,0.5)},font=\scriptsize,anchor=south}]
\addplot[color=paperOrange,mark=o,mark size=1.8pt,line width=1pt] coordinates {(1,3.079216312843) (2,3.049682619257) (3,3.047013672419) (4,3.047013672419) (5,3.047013672419)};
\addplot[color=paperGreen,mark=o,mark size=1.8pt,line width=1pt] coordinates {(1,1.929482167217) (2,1.783514500944) (3,1.752866287447) (4,1.751020400501) (5,1.750813189148)};
\nextgroupplot[ymin=1.5,ymax=4.5,ytick={1.5,2.0,2.5,3.0,3.5,4.0,4.5},scaled y ticks=false,yticklabel style={font=\scriptsize,/pgf/number format/fixed,/pgf/number format/fixed zerofill,/pgf/number format/precision=1}]
\addplot[color=paperOrange,mark=o,mark size=1.8pt,line width=1pt] coordinates {(1,4.245084865171) (2,4.148276855509) (3,4.136960813404) (4,4.136960813404) (5,4.136960813404)};
\addplot[color=paperGreen,mark=o,mark size=1.8pt,line width=1pt] coordinates {(1,2.806266430960) (2,2.709216511055) (3,2.695565132234) (4,2.693467102618) (5,2.692771463932)};
\addplot[color=paperPink,mark=o,mark size=1.8pt,line width=1pt] coordinates {(1,1.989628806744) (2,1.897649857742) (3,1.865953659479) (4,1.853759944161) (5,1.850743391209)};
\nextgroupplot[ymin=1.5,ymax=5.0,ytick={1.5,2.0,2.5,3.0,3.5,4.0,4.5,5.0},scaled y ticks=false,yticklabel style={font=\scriptsize,/pgf/number format/fixed,/pgf/number format/fixed zerofill,/pgf/number format/precision=1}]
\addplot[color=paperOrange,mark=o,mark size=1.8pt,line width=1pt] coordinates {(1,4.819121653911) (2,4.684663579190) (3,4.664740778249) (4,4.664740778249) (5,4.664740778249)};
\addplot[color=paperGreen,mark=o,mark size=1.8pt,line width=1pt] coordinates {(1,4.138499512226) (2,3.953056468522) (3,3.917900225879) (4,3.901405426568) (5,3.898691663901)};
\addplot[color=paperPink,mark=o,mark size=1.8pt,line width=1pt] coordinates {(1,2.790977355492) (2,2.662480762705) (3,2.642407386330) (4,2.635062747910) (5,2.627616605185)};
\addplot[color=paperBlue,mark=o,mark size=1.8pt,line width=1pt] coordinates {(1,2.104377474100) (2,1.993112257783) (3,1.955240985364) (4,1.944875021317) (5,1.937016766301)};
\end{groupplot}
\end{tikzpicture}
\caption{Cross-area}
\label{fig:search-level-crossarea}
\end{subfigure}
\caption{Sensitivity to search level $L$. Columns correspond to the $100$, $200$, and $400$\,m scales, which contain two, three, and four search levels, respectively; rows report mean and median top-$k$ oracle error. Results are reported for both same-area and cross-area evaluation at all three scales.}
\label{fig:search-level-full}
\end{figure}

Figures~\ref{fig:search-level-full}--\ref{fig:search-budget-full} extend the search analysis in Figure~\ref{fig:search-ablation} to all VIGOR-Pro scales under both same-area and cross-area evaluation.

\paragraph{Search level.}
Increasing the number of search levels becomes increasingly important as the aerial region expands. At $400$\,m and $k=2$, increasing $L$ from $1$ to $4$ reduces the mean oracle error from $12.49$ to $10.43$\,m on same-area and from $19.26$ to $15.64$\,m on cross-area; the corresponding median errors decrease from $3.05$ to $1.76$\,m and from $4.68$ to $1.99$\,m. The same trend holds at smaller scales, but the gains become more pronounced over broader regions, supporting the need for iterative zooming when the initial localization prior is coarse.

\paragraph{Sampling count.}
Increasing $M$ provides the largest gain from $8$ to $16$, with further sampling yielding diminishing improvements. At $400$\,m and $k=2$, the same-area mean error decreases from $12.94$ to $10.43$\,m and further to $9.54$\,m at $M=32$, while cross-area error decreases from $18.29$ to $15.64$\,m and $15.17$\,m, respectively. Median errors change much less and are not consistently improved by larger $M$, suggesting that additional samples primarily help recover difficult cases by exposing alternative plausible modes.

\paragraph{Pruning budget.}
A small budget of $B=2$ degrades search quality, particularly at larger scales, whereas $B=4$ and $8$ yield similar top-$2$ performance across both evaluation settings. Larger budgets become more beneficial mainly when considering deeper candidate lists. These results indicate that once enough competing branches are retained, search is comparatively insensitive to the exact pruning budget, supporting our default choice of $B=4$ for the compact top-$2$ shortlist used by the verifier.

\clearpage

\clearpage
\begin{figure}[p]
    \centering
    \begin{subfigure}[t]{0.98\textwidth}
        \centering
        \includegraphics[width=\linewidth]{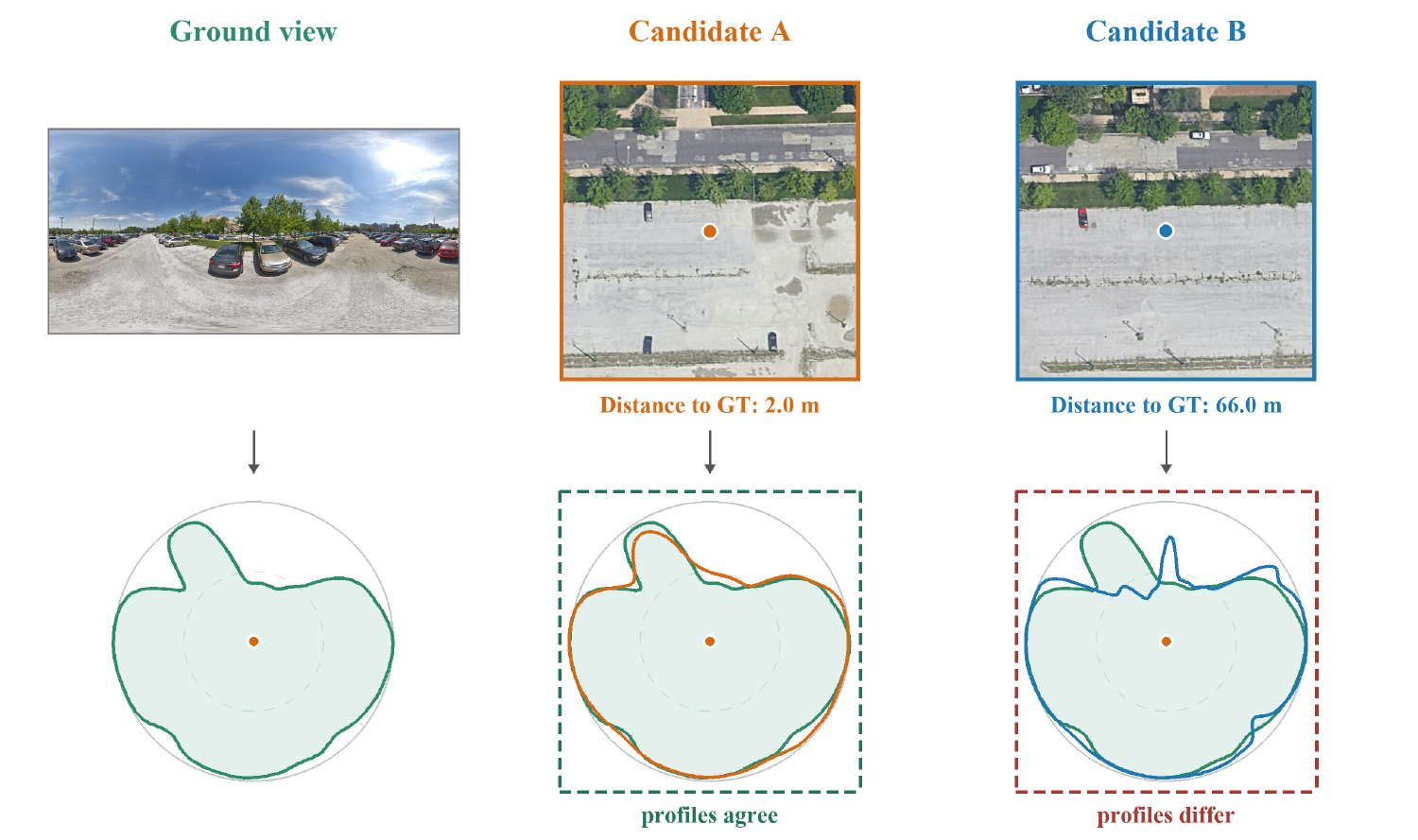}
        \caption{}
        \label{fig:depth_ring_cases:a}
    \end{subfigure}

    \vspace{1.6em}

    \begin{subfigure}[t]{0.98\textwidth}
        \centering
        \includegraphics[width=\linewidth]{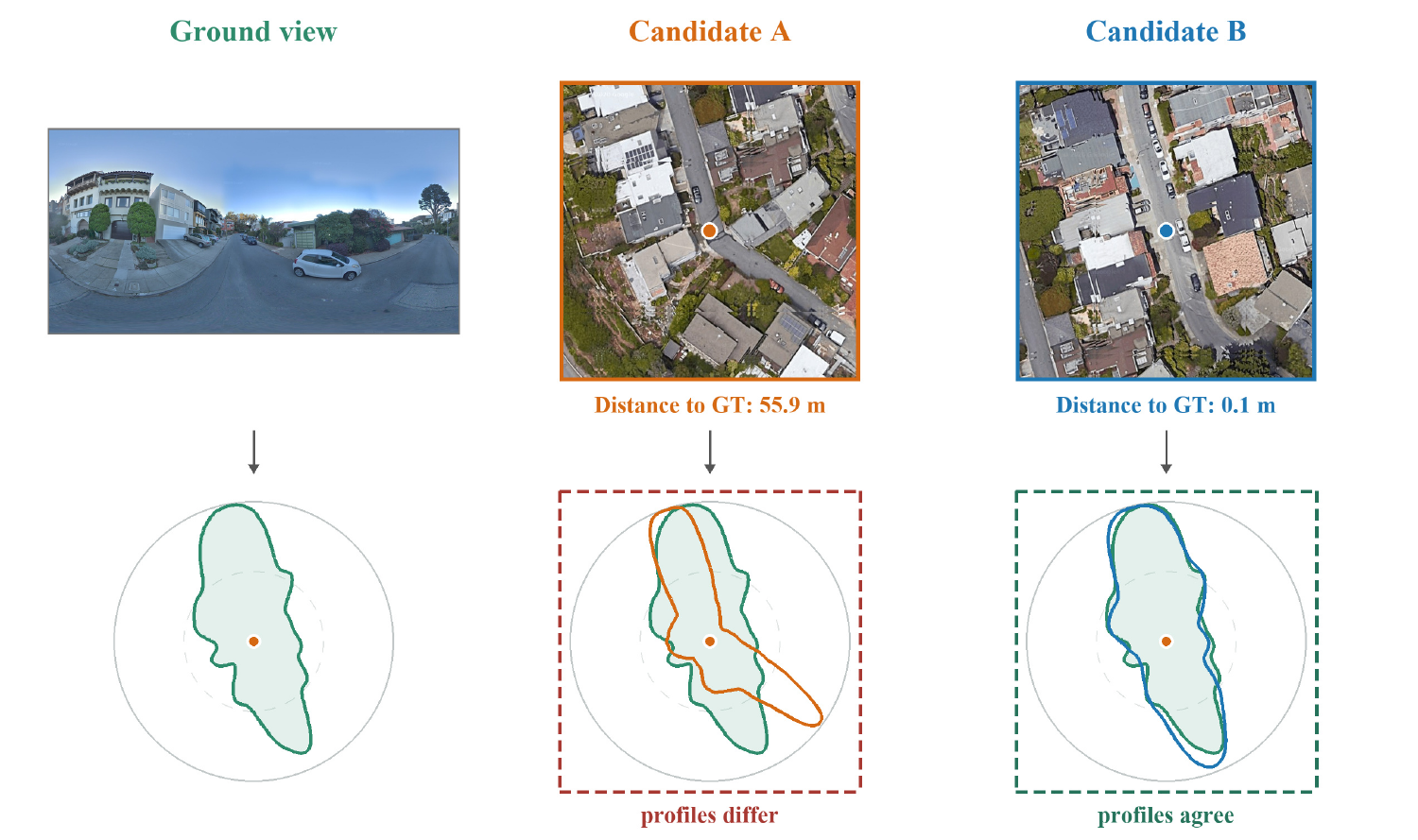}
        \caption{}
        \label{fig:depth_ring_cases:b}
    \end{subfigure}

\caption{Qualitative examples of depth rings for cross-view matching.
Each panel compares the ground-view depth ring with predicted aerial depth rings at two candidate locations. Candidates closer to the ground-truth camera location exhibit stronger correspondence in surrounding spatial layout, while incorrect candidates show clear geometric discrepancies.}

    \label{fig:depth_ring_cases}
\end{figure}

\clearpage

\begin{figure}[p]
    \ContinuedFloat
    \centering
    \setcounter{subfigure}{2}
    \begin{subfigure}[t]{0.98\textwidth}
        \centering
        \includegraphics[width=\linewidth]{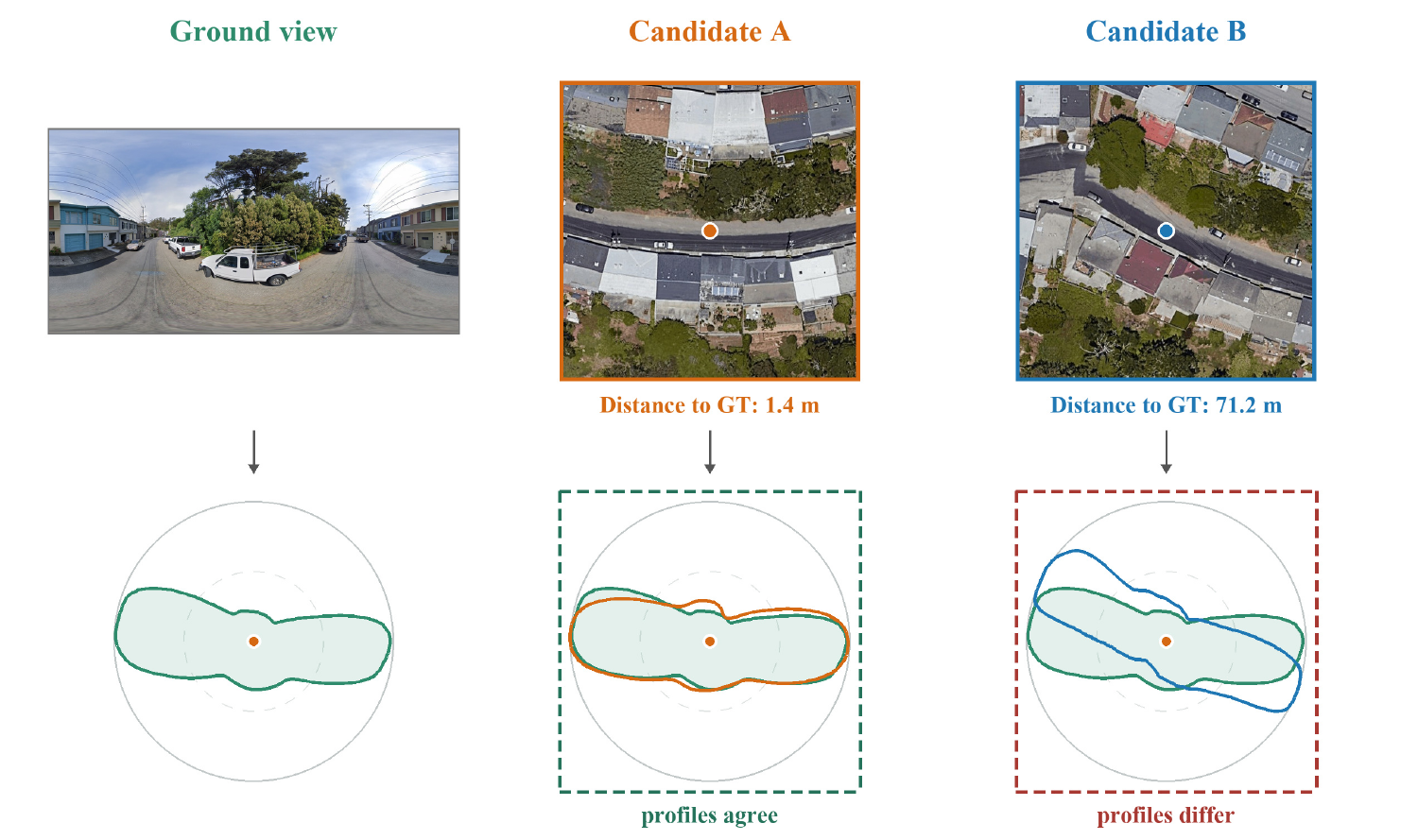}
        \caption{}
        \label{fig:depth_ring_cases:c}
    \end{subfigure}

    \vspace{1.6em}

    \begin{subfigure}[t]{0.98\textwidth}
        \centering
        \includegraphics[width=\linewidth]{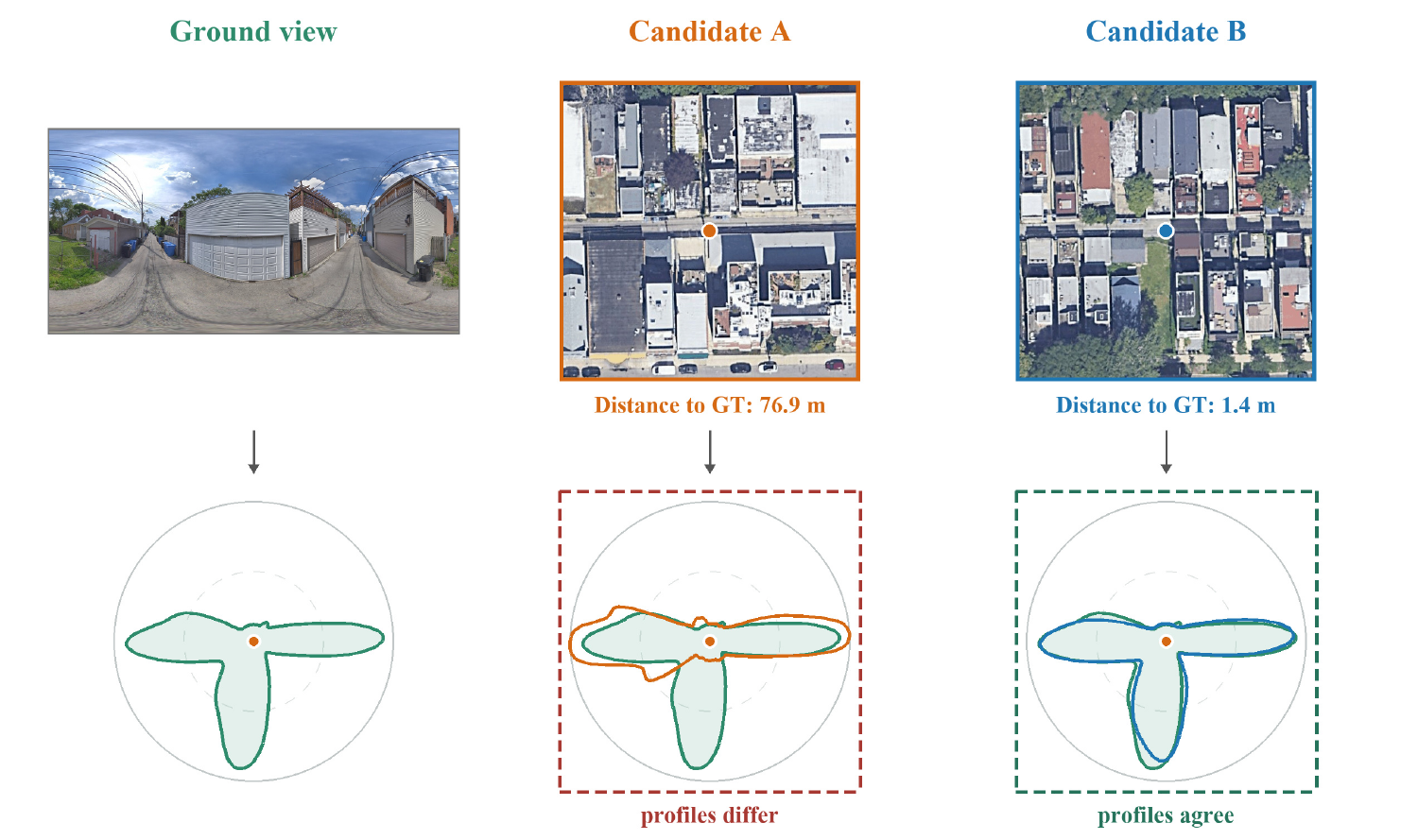}
        \caption{}
        \label{fig:depth_ring_cases:d}
    \end{subfigure}

\caption{Qualitative examples of depth rings for cross-view matching (continued).}
\end{figure}

\clearpage

\clearpage
\begin{figure}[!ht]
\centering
\pgfplotslegendfromname{searchsampleslegend}
\vspace{0.08cm}
\begin{subfigure}[t]{\linewidth}
\centering
\hspace*{-0.6cm}
\begin{tikzpicture}
\begin{groupplot}[searchaxis,width=5.25cm,height=5.25cm,label style={font=\scriptsize},tick label style={font=\scriptsize},group style={group size=3 by 2,horizontal sep=0.7cm,vertical sep=0.55cm}]
\nextgroupplot[ymin=2.5,ymax=4.5,ytick={2.5,3,3.5,4,4.5},title={$100$\,m},xlabel={},ylabel={Mean Distance Error (m) $\downarrow$},y label style={at={(axis description cs:-0.125,0.5)},font=\scriptsize,anchor=south}]
\addplot[color=paperOrange,mark=o,mark size=1.8pt,line width=1pt] coordinates {(1,4.368302020264) (2,3.746124470886) (3,3.693845957727) (4,3.689279268590) (5,3.687822432880)};
\addplot[color=paperGreen,mark=o,mark size=1.8pt,line width=1pt] coordinates {(1,4.32) (2,3.372402373300) (3,3.197918496108) (4,3.159989079638) (5,3.141394495653)};
\addplot[color=paperPink,mark=o,mark size=1.8pt,line width=1pt] coordinates {(1,4.310877607004) (2,3.279357251492) (3,3.084667267277) (4,3.024502109180) (5,2.987412699012)};
\addplot[color=paperBlue,mark=o,mark size=1.8pt,line width=1pt] coordinates {(1,4.326553500148) (2,3.226542507496) (3,3.058588036526) (4,2.976041119990) (5,2.937459600211)};
\nextgroupplot[ymin=4,ymax=9,ytick={4,5,6,7,8,9},title={$200$\,m},xlabel={}]
\addplot[color=paperOrange,mark=o,mark size=1.8pt,line width=1pt] coordinates {(1,8.490321144907) (2,6.894474491915) (3,6.661416191111) (4,6.612754894351) (5,6.595555587929)};
\addplot[color=paperGreen,mark=o,mark size=1.8pt,line width=1pt] coordinates {(1,8.03) (2,5.691253085257) (3,5.270550943366) (4,5.123494151369) (5,5.069070661640)};
\addplot[color=paperPink,mark=o,mark size=1.8pt,line width=1pt] coordinates {(1,8.196583697578) (2,5.423463151844) (3,4.911096321232) (4,4.689190340843) (5,4.612779830471)};
\addplot[color=paperBlue,mark=o,mark size=1.8pt,line width=1pt] coordinates {(1,7.964701407813) (2,5.273768871665) (3,4.750403671700) (4,4.575326668312) (5,4.499513407539)};
\nextgroupplot[ymin=7,ymax=19,ytick={7,9,11,13,15,17,19},title={$400$\,m},xlabel={},legend to name=searchsampleslegend,legend columns=-1,legend style={font=\small,fill=none,draw=none,inner sep=0pt,legend columns=-1,/tikz/every even column/.append style={column sep=0.45cm}}]
\addplot[color=paperOrange,mark=o,mark size=1.8pt,line width=1pt] coordinates {(1,16.881748198860) (2,12.943944137053) (3,12.188305442047) (4,11.904154295672) (5,11.827704707612)};
\addlegendentry{$M=8$}
\addplot[color=paperGreen,mark=o,mark size=1.8pt,line width=1pt] coordinates {(1,15.60) (2,10.426106546195) (3,9.124446608028) (4,8.632364522232) (5,8.521661382657)};
\addlegendentry{$M=16$}
\addplot[color=paperPink,mark=o,mark size=1.8pt,line width=1pt] coordinates {(1,15.545560407746) (2,9.765733873490) (3,8.367693861338) (4,7.945173221986) (5,7.840255849653)};
\addlegendentry{$M=24$}
\addplot[color=paperBlue,mark=o,mark size=1.8pt,line width=1pt] coordinates {(1,15.387209987713) (2,9.543130732992) (3,8.025855081192) (4,7.569422040370) (5,7.442849424495)};
\addlegendentry{$M=32$}
\nextgroupplot[ymin=1.5,ymax=1.8,ytick={1.5,1.6,1.7,1.8},scaled y ticks=false,yticklabel style={font=\scriptsize,/pgf/number format/fixed,/pgf/number format/fixed zerofill,/pgf/number format/precision=1},ylabel={Median Distance Error (m) $\downarrow$},y label style={at={(axis description cs:-0.125,0.5)},font=\scriptsize,anchor=south}]
\addplot[color=paperOrange,mark=o,mark size=1.8pt,line width=1pt] coordinates {(1,1.700300135238) (2,1.617040075508) (3,1.611659400673) (4,1.610217793635) (5,1.609200257481)};
\addplot[color=paperGreen,mark=o,mark size=1.8pt,line width=1pt] coordinates {(1,1.700046448810) (2,1.595970154412) (3,1.569001726999) (4,1.566659371036) (5,1.566494691972)};
\addplot[color=paperPink,mark=o,mark size=1.8pt,line width=1pt] coordinates {(1,1.705030666259) (2,1.615640260350) (3,1.592583332618) (4,1.588221531137) (5,1.587019095327)};
\addplot[color=paperBlue,mark=o,mark size=1.8pt,line width=1pt] coordinates {(1,1.732626407891) (2,1.642561041378) (3,1.621121479680) (4,1.618613237773) (5,1.617705354476)};
\nextgroupplot[ymin=1.6,ymax=1.9,ytick={1.6,1.7,1.8,1.9},scaled y ticks=false,yticklabel style={font=\scriptsize,/pgf/number format/fixed,/pgf/number format/fixed zerofill,/pgf/number format/precision=1}]
\addplot[color=paperOrange,mark=o,mark size=1.8pt,line width=1pt] coordinates {(1,1.843397917073) (2,1.744658785950) (3,1.731494118714) (4,1.727898689286) (5,1.727898689286)};
\addplot[color=paperGreen,mark=o,mark size=1.8pt,line width=1pt] coordinates {(1,1.860633237955) (2,1.714787141381) (3,1.655824351188) (4,1.651750108028) (5,1.651500397897)};
\addplot[color=paperPink,mark=o,mark size=1.8pt,line width=1pt] coordinates {(1,1.856945564663) (2,1.706086512194) (3,1.662534698200) (4,1.658119165721) (5,1.654311808778)};
\addplot[color=paperBlue,mark=o,mark size=1.8pt,line width=1pt] coordinates {(1,1.876810327561) (2,1.758526834228) (3,1.730613901992) (4,1.723307321128) (5,1.721390518719)};
\nextgroupplot[ymin=1.6,ymax=2.0,ytick={1.6,1.7,1.8,1.9,2.0},scaled y ticks=false,yticklabel style={font=\scriptsize,/pgf/number format/fixed,/pgf/number format/fixed zerofill,/pgf/number format/precision=1}]
\addplot[color=paperOrange,mark=o,mark size=1.8pt,line width=1pt] coordinates {(1,1.954088068698) (2,1.827477993042) (3,1.805588479061) (4,1.802672200986) (5,1.801619161213)};
\addplot[color=paperGreen,mark=o,mark size=1.8pt,line width=1pt] coordinates {(1,1.927912952708) (2,1.759645780394) (3,1.702350313192) (4,1.695391442547) (5,1.690386071844)};
\addplot[color=paperPink,mark=o,mark size=1.8pt,line width=1pt] coordinates {(1,1.951546125435) (2,1.778677689860) (3,1.717673334701) (4,1.705406816629) (5,1.700685986210)};
\addplot[color=paperBlue,mark=o,mark size=1.8pt,line width=1pt] coordinates {(1,1.968500961827) (2,1.802171782867) (3,1.762662773584) (4,1.754363611590) (5,1.749903789315)};
\end{groupplot}
\end{tikzpicture}
\caption{Same-area}
\label{fig:search-samples-samearea}
\end{subfigure}
\vspace{0.25cm}
\begin{subfigure}[t]{\linewidth}
\centering
\hspace*{-0.6cm}
\begin{tikzpicture}
\begin{groupplot}[searchaxis,width=5.25cm,height=5.25cm,label style={font=\scriptsize},tick label style={font=\scriptsize},group style={group size=3 by 2,horizontal sep=0.7cm,vertical sep=0.55cm}]
\nextgroupplot[ymin=3,ymax=5,ytick={3,3.5,4,4.5,5},title={$100$\,m},xlabel={},ylabel={Mean Distance Error (m) $\downarrow$},y label style={at={(axis description cs:-0.125,0.5)},font=\scriptsize,anchor=south}]
\addplot[color=paperOrange,mark=o,mark size=1.8pt,line width=1pt] coordinates {(1,4.916051902137) (2,4.166728126539) (3,4.065299802574) (4,4.054885442907) (5,4.053059266384)};
\addplot[color=paperGreen,mark=o,mark size=1.8pt,line width=1pt] coordinates {(1,4.95) (2,3.769998105402) (3,3.498888877936) (4,3.462195412388) (5,3.449691138383)};
\addplot[color=paperPink,mark=o,mark size=1.8pt,line width=1pt] coordinates {(1,4.821677961838) (2,3.690972711339) (3,3.409100935148) (4,3.349075823978) (5,3.318779342663)};
\addplot[color=paperBlue,mark=o,mark size=1.8pt,line width=1pt] coordinates {(1,4.853846390350) (2,3.708232787157) (3,3.445948001892) (4,3.362508941090) (5,3.324502437356)};
\nextgroupplot[ymin=5,ymax=12,ytick={5,6,7,8,9,10,11,12},title={$200$\,m},xlabel={}]
\addplot[color=paperOrange,mark=o,mark size=1.8pt,line width=1pt] coordinates {(1,11.111662243704) (2,8.714585500800) (3,8.331129473112) (4,8.239576898136) (5,8.231028215670)};
\addplot[color=paperGreen,mark=o,mark size=1.8pt,line width=1pt] coordinates {(1,11.01) (2,7.638748511999) (3,6.840987895277) (4,6.631419270730) (5,6.537599874119)};
\addplot[color=paperPink,mark=o,mark size=1.8pt,line width=1pt] coordinates {(1,11.182894648173) (2,7.376399932911) (3,6.423052193633) (4,6.117803642830) (5,5.991702184847)};
\addplot[color=paperBlue,mark=o,mark size=1.8pt,line width=1pt] coordinates {(1,11.173564477262) (2,7.294428798515) (3,6.373391472216) (4,6.024120420637) (5,5.931643534839)};
\nextgroupplot[ymin=10,ymax=25,ytick={10,13,16,19,22,25},title={$400$\,m},xlabel={}]
\addplot[color=paperOrange,mark=o,mark size=1.8pt,line width=1pt] coordinates {(1,23.642087882056) (2,18.285163334803) (3,17.038210287152) (4,16.665010767326) (5,16.578182875721)};
\addplot[color=paperGreen,mark=o,mark size=1.8pt,line width=1pt] coordinates {(1,23.99) (2,15.637527531406) (3,13.628519284830) (4,12.883404732705) (5,12.539866095711)};
\addplot[color=paperPink,mark=o,mark size=1.8pt,line width=1pt] coordinates {(1,23.104290442192) (2,15.399484805439) (3,12.868123849538) (4,11.764331570785) (5,11.405563700056)};
\addplot[color=paperBlue,mark=o,mark size=1.8pt,line width=1pt] coordinates {(1,23.242082428914) (2,15.170009335359) (3,12.360771595259) (4,11.366013387420) (5,11.099575043568)};
\nextgroupplot[ymin=1.7,ymax=2.1,ytick={1.7,1.8,1.9,2.0,2.1},scaled y ticks=false,yticklabel style={font=\scriptsize,/pgf/number format/fixed,/pgf/number format/fixed zerofill,/pgf/number format/precision=1},ylabel={Median Distance Error (m) $\downarrow$},y label style={at={(axis description cs:-0.125,0.5)},font=\scriptsize,anchor=south}]
\addplot[color=paperOrange,mark=o,mark size=1.8pt,line width=1pt] coordinates {(1,1.838892512793) (2,1.761876707817) (3,1.752655404427) (4,1.750516294845) (5,1.750192591070)};
\addplot[color=paperGreen,mark=o,mark size=1.8pt,line width=1pt] coordinates {(1,1.929482167217) (2,1.783514500944) (3,1.752866287447) (4,1.751020400501) (5,1.750813189148)};
\addplot[color=paperPink,mark=o,mark size=1.8pt,line width=1pt] coordinates {(1,1.919270795287) (2,1.806536780671) (3,1.780082964809) (4,1.776589033899) (5,1.774176903996)};
\addplot[color=paperBlue,mark=o,mark size=1.8pt,line width=1pt] coordinates {(1,2.041929368873) (2,1.934010054022) (3,1.904813822913) (4,1.900275451391) (5,1.899884630167)};
\nextgroupplot[ymin=1.8,ymax=2.2,ytick={1.8,1.9,2.0,2.1,2.2},scaled y ticks=false,yticklabel style={font=\scriptsize,/pgf/number format/fixed,/pgf/number format/fixed zerofill,/pgf/number format/precision=1}]
\addplot[color=paperOrange,mark=o,mark size=1.8pt,line width=1pt] coordinates {(1,1.916342087491) (2,1.837905335166) (3,1.825171241888) (4,1.823391534481) (5,1.822687629066)};
\addplot[color=paperGreen,mark=o,mark size=1.8pt,line width=1pt] coordinates {(1,1.989628806744) (2,1.897649857742) (3,1.865953659479) (4,1.853759944161) (5,1.850743391209)};
\addplot[color=paperPink,mark=o,mark size=1.8pt,line width=1pt] coordinates {(1,2.017491102194) (2,1.918495017001) (3,1.882579985379) (4,1.874617782147) (5,1.870605438743)};
\addplot[color=paperBlue,mark=o,mark size=1.8pt,line width=1pt] coordinates {(1,2.159996545401) (2,2.069014233077) (3,2.043621489848) (4,2.022137734343) (5,2.018980735380)};
\nextgroupplot[ymin=1.8,ymax=2.3,ytick={1.8,1.9,2.0,2.1,2.2,2.3},scaled y ticks=false,yticklabel style={font=\scriptsize,/pgf/number format/fixed,/pgf/number format/fixed zerofill,/pgf/number format/precision=1}]
\addplot[color=paperOrange,mark=o,mark size=1.8pt,line width=1pt] coordinates {(1,1.984819330515) (2,1.909672246117) (3,1.893707439283) (4,1.886510412269) (5,1.886485313996)};
\addplot[color=paperGreen,mark=o,mark size=1.8pt,line width=1pt] coordinates {(1,2.104377474100) (2,1.993112257783) (3,1.955240985364) (4,1.944875021317) (5,1.937016766301)};
\addplot[color=paperPink,mark=o,mark size=1.8pt,line width=1pt] coordinates {(1,2.101307355536) (2,1.994538272614) (3,1.953305543423) (4,1.940389187040) (5,1.935278480321)};
\addplot[color=paperBlue,mark=o,mark size=1.8pt,line width=1pt] coordinates {(1,2.269244433555) (2,2.148672574618) (3,2.100837528419) (4,2.086602445008) (5,2.080978533368)};
\end{groupplot}
\end{tikzpicture}
\caption{Cross-area}
\label{fig:search-samples-crossarea}
\end{subfigure}
\caption{Sensitivity to sampling count $M$. Columns correspond to the $100$, $200$, and $400$\,m scales; rows report mean and median top-$k$ oracle error. Results are reported for both same-area and cross-area evaluation at all three scales.}
\label{fig:search-samples-full}
\end{figure}

\clearpage
\begin{figure}[!ht]
\centering
\pgfplotslegendfromname{searchbudgetlegend}
\vspace{0.08cm}
\begin{subfigure}[t]{\linewidth}
\centering
\hspace*{-0.6cm}
\begin{tikzpicture}
\begin{groupplot}[searchaxis,width=5.25cm,height=5.25cm,label style={font=\scriptsize},tick label style={font=\scriptsize},group style={group size=3 by 2,horizontal sep=0.7cm,vertical sep=0.55cm}]
\nextgroupplot[ymin=3,ymax=4.5,ytick={3,3.5,4,4.5},title={$100$\,m},xlabel={},ylabel={Mean Distance Error (m) $\downarrow$},y label style={at={(axis description cs:-0.125,0.5)},font=\scriptsize,anchor=south}]
\addplot[color=paperOrange,mark=o,mark size=1.8pt,line width=1pt] coordinates {(1,4.376012399947) (2,3.402934549375) (3,3.252445463781) (4,3.215575260298) (5,3.208287085777)};
\addplot[color=paperGreen,mark=o,mark size=1.8pt,line width=1pt] coordinates {(1,4.380302545979) (2,3.381985903285) (3,3.213080250320) (4,3.154820737923) (5,3.138409330435)};
\addplot[color=paperPink,mark=o,mark size=1.8pt,line width=1pt] coordinates {(1,4.32) (2,3.372402373300) (3,3.197918496108) (4,3.159989079638) (5,3.141394495653)};
\addplot[color=paperBlue,mark=o,mark size=1.8pt,line width=1pt] coordinates {(1,4.357020739479) (2,3.369130095169) (3,3.210964674090) (4,3.143457661511) (5,3.128691615865)};
\nextgroupplot[ymin=4,ymax=9,ytick={4,5,6,7,8,9},title={$200$\,m},xlabel={}]
\addplot[color=paperOrange,mark=o,mark size=1.8pt,line width=1pt] coordinates {(1,8.357281800087) (2,5.843456203763) (3,5.663993401253) (4,5.610598714474) (5,5.604939545032)};
\addplot[color=paperGreen,mark=o,mark size=1.8pt,line width=1pt] coordinates {(1,8.405111405428) (2,5.711281135664) (3,5.263594556711) (4,5.166099034446) (5,5.133796018202)};
\addplot[color=paperPink,mark=o,mark size=1.8pt,line width=1pt] coordinates {(1,8.03) (2,5.691253085257) (3,5.270550943366) (4,5.123494151369) (5,5.069070661640)};
\addplot[color=paperBlue,mark=o,mark size=1.8pt,line width=1pt] coordinates {(1,8.248127162059) (2,5.702619992791) (3,5.232487144148) (4,5.071687483721) (5,5.017275396341)};
\nextgroupplot[ymin=8,ymax=18,ytick={8,10,12,14,16,18},title={$400$\,m},xlabel={},legend to name=searchbudgetlegend,legend columns=-1,legend style={font=\small,fill=none,draw=none,inner sep=0pt,legend columns=-1,/tikz/every even column/.append style={column sep=0.45cm}}]
\addplot[color=paperOrange,mark=o,mark size=1.8pt,line width=1pt] coordinates {(1,16.041563471389) (2,11.361199401392) (3,11.059704542759) (4,10.916371718046) (5,10.906736501521)};
\addlegendentry{$B=2$}
\addplot[color=paperGreen,mark=o,mark size=1.8pt,line width=1pt] coordinates {(1,15.708941023092) (2,10.348947208346) (3,9.428837729437) (4,9.207351676484) (5,9.120443992991)};
\addlegendentry{$B=3$}
\addplot[color=paperPink,mark=o,mark size=1.8pt,line width=1pt] coordinates {(1,15.60) (2,10.426106546195) (3,9.124446608028) (4,8.632364522232) (5,8.521661382657)};
\addlegendentry{$B=4$}
\addplot[color=paperBlue,mark=o,mark size=1.8pt,line width=1pt] coordinates {(1,16.093198292324) (2,10.456816468357) (3,9.042126539025) (4,8.520012043208) (5,8.335077525865)};
\addlegendentry{$B=8$}
\nextgroupplot[ymin=1.5,ymax=1.8,ytick={1.5,1.6,1.7,1.8},scaled y ticks=false,yticklabel style={font=\scriptsize,/pgf/number format/fixed,/pgf/number format/fixed zerofill,/pgf/number format/precision=1},ylabel={Median Distance Error (m) $\downarrow$},y label style={at={(axis description cs:-0.125,0.5)},font=\scriptsize,anchor=south}]
\addplot[color=paperOrange,mark=o,mark size=1.8pt,line width=1pt] coordinates {(1,1.692119753010) (2,1.599779564539) (3,1.570480955011) (4,1.568648386085) (5,1.568520121350)};
\addplot[color=paperGreen,mark=o,mark size=1.8pt,line width=1pt] coordinates {(1,1.694591670191) (2,1.592095561478) (3,1.566861292882) (4,1.562955301515) (5,1.561605786734)};
\addplot[color=paperPink,mark=o,mark size=1.8pt,line width=1pt] coordinates {(1,1.700046448810) (2,1.595970154412) (3,1.569001726999) (4,1.566659371036) (5,1.566494691972)};
\addplot[color=paperBlue,mark=o,mark size=1.8pt,line width=1pt] coordinates {(1,1.697052509293) (2,1.598007231233) (3,1.572033251085) (4,1.570192643708) (5,1.569358816891)};
\nextgroupplot[ymin=1.6,ymax=1.9,ytick={1.6,1.7,1.8,1.9},scaled y ticks=false,yticklabel style={font=\scriptsize,/pgf/number format/fixed,/pgf/number format/fixed zerofill,/pgf/number format/precision=1}]
\addplot[color=paperOrange,mark=o,mark size=1.8pt,line width=1pt] coordinates {(1,1.870644741863) (2,1.707993190669) (3,1.653353422049) (4,1.649379995623) (5,1.648854853796)};
\addplot[color=paperGreen,mark=o,mark size=1.8pt,line width=1pt] coordinates {(1,1.853454774713) (2,1.707394470355) (3,1.654285917050) (4,1.650885015756) (5,1.649804304882)};
\addplot[color=paperPink,mark=o,mark size=1.8pt,line width=1pt] coordinates {(1,1.860633237955) (2,1.714787141381) (3,1.655824351188) (4,1.651750108028) (5,1.651500397897)};
\addplot[color=paperBlue,mark=o,mark size=1.8pt,line width=1pt] coordinates {(1,1.848400551589) (2,1.696749026609) (3,1.651382755977) (4,1.644620468013) (5,1.643019167805)};
\nextgroupplot[ymin=1.6,ymax=2.0,ytick={1.6,1.7,1.8,1.9,2.0},scaled y ticks=false,yticklabel style={font=\scriptsize,/pgf/number format/fixed,/pgf/number format/fixed zerofill,/pgf/number format/precision=1}]
\addplot[color=paperOrange,mark=o,mark size=1.8pt,line width=1pt] coordinates {(1,1.920989137415) (2,1.752744101778) (3,1.698283809772) (4,1.695564865114) (5,1.694712532372)};
\addplot[color=paperGreen,mark=o,mark size=1.8pt,line width=1pt] coordinates {(1,1.935036497406) (2,1.749914640117) (3,1.691365450668) (4,1.681341876917) (5,1.676646536260)};
\addplot[color=paperPink,mark=o,mark size=1.8pt,line width=1pt] coordinates {(1,1.927912952708) (2,1.759645780394) (3,1.702350313192) (4,1.695391442547) (5,1.690386071844)};
\addplot[color=paperBlue,mark=o,mark size=1.8pt,line width=1pt] coordinates {(1,1.934428108102) (2,1.756988472415) (3,1.692908408208) (4,1.680394053621) (5,1.678352082124)};
\end{groupplot}
\end{tikzpicture}
\caption{Same-area}
\label{fig:search-budget-samearea}
\end{subfigure}
\vspace{0.25cm}
\begin{subfigure}[t]{\linewidth}
\centering
\hspace*{-0.6cm}
\begin{tikzpicture}
\begin{groupplot}[searchaxis,width=5.25cm,height=5.25cm,label style={font=\scriptsize},tick label style={font=\scriptsize},group style={group size=3 by 2,horizontal sep=0.7cm,vertical sep=0.55cm}]
\nextgroupplot[ymin=3,ymax=5,ytick={3,3.5,4,4.5,5},title={$100$\,m},xlabel={},ylabel={Mean Distance Error (m) $\downarrow$},y label style={at={(axis description cs:-0.125,0.5)},font=\scriptsize,anchor=south}]
\addplot[color=paperOrange,mark=o,mark size=1.8pt,line width=1pt] coordinates {(1,4.856530734112) (2,3.782473931749) (3,3.520159479202) (4,3.491892422682) (5,3.483041444559)};
\addplot[color=paperGreen,mark=o,mark size=1.8pt,line width=1pt] coordinates {(1,4.905933859066) (2,3.772741448965) (3,3.490590507043) (4,3.446463221224) (5,3.426001909798)};
\addplot[color=paperPink,mark=o,mark size=1.8pt,line width=1pt] coordinates {(1,4.95) (2,3.769998105402) (3,3.498888877936) (4,3.462195412388) (5,3.449691138383)};
\addplot[color=paperBlue,mark=o,mark size=1.8pt,line width=1pt] coordinates {(1,4.848839953389) (2,3.749278311920) (3,3.479982882477) (4,3.450807753090) (5,3.433641705211)};
\nextgroupplot[ymin=6,ymax=12,ytick={6,7,8,9,10,11,12},title={$200$\,m},xlabel={}]
\addplot[color=paperOrange,mark=o,mark size=1.8pt,line width=1pt] coordinates {(1,11.093563669079) (2,7.935495374877) (3,7.485972186083) (4,7.354935703448) (5,7.336776256045)};
\addplot[color=paperGreen,mark=o,mark size=1.8pt,line width=1pt] coordinates {(1,11.054318649327) (2,7.727325173836) (3,6.902158948985) (4,6.729895107336) (5,6.665648606339)};
\addplot[color=paperPink,mark=o,mark size=1.8pt,line width=1pt] coordinates {(1,11.01) (2,7.638748511999) (3,6.840987895277) (4,6.631419270730) (5,6.537599874119)};
\addplot[color=paperBlue,mark=o,mark size=1.8pt,line width=1pt] coordinates {(1,11.075723037405) (2,7.768042689846) (3,6.934582617397) (4,6.726343126519) (5,6.641787264851)};
\nextgroupplot[ymin=11,ymax=25,ytick={11,13,15,17,19,21,23,25},title={$400$\,m},xlabel={}]
\addplot[color=paperOrange,mark=o,mark size=1.8pt,line width=1pt] coordinates {(1,24.202915320147) (2,17.366227785985) (3,16.296290485886) (4,15.946072693608) (5,15.912935061606)};
\addplot[color=paperGreen,mark=o,mark size=1.8pt,line width=1pt] coordinates {(1,23.171612167828) (2,16.033830725422) (3,13.932087095743) (4,13.391599732453) (5,13.239266874582)};
\addplot[color=paperPink,mark=o,mark size=1.8pt,line width=1pt] coordinates {(1,23.99) (2,15.637527531406) (3,13.628519284830) (4,12.883404732705) (5,12.539866095711)};
\addplot[color=paperBlue,mark=o,mark size=1.8pt,line width=1pt] coordinates {(1,23.238161407525) (2,15.611599877870) (3,13.259382808929) (4,12.545851946475) (5,12.128061577989)};
\nextgroupplot[ymin=1.7,ymax=2.0,ytick={1.7,1.8,1.9,2.0},scaled y ticks=false,yticklabel style={font=\scriptsize,/pgf/number format/fixed,/pgf/number format/fixed zerofill,/pgf/number format/precision=1},ylabel={Median Distance Error (m) $\downarrow$},y label style={at={(axis description cs:-0.125,0.5)},font=\scriptsize,anchor=south}]
\addplot[color=paperOrange,mark=o,mark size=1.8pt,line width=1pt] coordinates {(1,1.925759223152) (2,1.777973036404) (3,1.743530386805) (4,1.741220591614) (5,1.740954410420)};
\addplot[color=paperGreen,mark=o,mark size=1.8pt,line width=1pt] coordinates {(1,1.920811273625) (2,1.780159821698) (3,1.744204908643) (4,1.739015563345) (5,1.738363881606)};
\addplot[color=paperPink,mark=o,mark size=1.8pt,line width=1pt] coordinates {(1,1.929482167217) (2,1.783514500944) (3,1.752866287447) (4,1.751020400501) (5,1.750813189148)};
\addplot[color=paperBlue,mark=o,mark size=1.8pt,line width=1pt] coordinates {(1,1.924021560658) (2,1.778871234703) (3,1.746488858201) (4,1.743495745365) (5,1.742436492349)};
\nextgroupplot[ymin=1.8,ymax=2.1,ytick={1.8,1.9,2.0,2.1},scaled y ticks=false,yticklabel style={font=\scriptsize,/pgf/number format/fixed,/pgf/number format/fixed zerofill,/pgf/number format/precision=1}]
\addplot[color=paperOrange,mark=o,mark size=1.8pt,line width=1pt] coordinates {(1,1.997776695355) (2,1.896430020043) (3,1.878205939854) (4,1.871818118812) (5,1.871252516714)};
\addplot[color=paperGreen,mark=o,mark size=1.8pt,line width=1pt] coordinates {(1,1.987428632138) (2,1.884726329684) (3,1.859189570214) (4,1.849703413200) (5,1.845794758357)};
\addplot[color=paperPink,mark=o,mark size=1.8pt,line width=1pt] coordinates {(1,1.989628806744) (2,1.897649857742) (3,1.865953659479) (4,1.853759944161) (5,1.850743391209)};
\addplot[color=paperBlue,mark=o,mark size=1.8pt,line width=1pt] coordinates {(1,1.992494700946) (2,1.901370604781) (3,1.874995540616) (4,1.865776614073) (5,1.862048166156)};
\nextgroupplot[ymin=1.9,ymax=2.2,ytick={1.9,2.0,2.1,2.2},scaled y ticks=false,yticklabel style={font=\scriptsize,/pgf/number format/fixed,/pgf/number format/fixed zerofill,/pgf/number format/precision=1}]
\addplot[color=paperOrange,mark=o,mark size=1.8pt,line width=1pt] coordinates {(1,2.105840369399) (2,2.007012167103) (3,1.985148612754) (4,1.981197988387) (5,1.980031873838)};
\addplot[color=paperGreen,mark=o,mark size=1.8pt,line width=1pt] coordinates {(1,2.109825577578) (2,1.993771528111) (3,1.953114983229) (4,1.946752692976) (5,1.942591703750)};
\addplot[color=paperPink,mark=o,mark size=1.8pt,line width=1pt] coordinates {(1,2.104377474100) (2,1.993112257783) (3,1.955240985364) (4,1.944875021317) (5,1.937016766301)};
\addplot[color=paperBlue,mark=o,mark size=1.8pt,line width=1pt] coordinates {(1,2.084383240365) (2,1.977007579133) (3,1.940961759710) (4,1.929529733166) (5,1.921158153601)};
\end{groupplot}
\end{tikzpicture}
\caption{Cross-area}
\label{fig:search-budget-crossarea}
\end{subfigure}
\caption{Sensitivity to pruning budget $B$. Columns correspond to the $100$, $200$, and $400$\,m scales; rows report mean and median top-$k$ oracle error. Results are reported for both same-area and cross-area evaluation at all three scales.}
\label{fig:search-budget-full}
\end{figure}

\clearpage
\newlength{\verifierringpromptwidth}
\begin{figure}[p]
\centering
\setlength{\fboxsep}{8pt}
\settowidth{\verifierringpromptwidth}{\scriptsize\texttt{- Image 1: a ground-level $360^\circ$ panoramic view from the unknown camera location.}}
\fbox{%
\begin{minipage}{\verifierringpromptwidth}
\scriptsize\raggedright
\textbf{User}\par\smallskip
\texttt{<image><image><image><image><image><image>}\par
\texttt{You are given six images:}\par
\texttt{- Image 1: a ground-level $360^\circ$ panoramic view from the unknown camera location.}\par
\texttt{- Image 2: the full satellite image with two candidate camera locations marked.}\par
\texttt{- Image 3: a $640\times640$ satellite crop centered on Candidate A (red marker).}\par
\texttt{- Image 4: a $640\times640$ satellite crop centered on Candidate B (blue marker).}\par
\texttt{- Image 5: a polar plot comparing two depth rings:}\par
\hspace*{1em}\texttt{* BLACK --- depth ring derived from Image 1 (the panorama).}\par
\hspace*{1em}\texttt{* RED --- depth ring predicted for Candidate A's satellite crop.}\par
\texttt{- Image 6: the same comparison for Candidate B:}\par
\hspace*{1em}\texttt{* BLACK --- the same panorama depth ring as in Image 5.}\par
\hspace*{1em}\texttt{* BLUE --- depth ring predicted for Candidate B's satellite crop.}\par
\medskip
\texttt{Candidate positions (normalized to [0, 1000]):}\par
\texttt{- A (Red): [$x_A$, $y_A$]}\par
\texttt{- B (Blue): [$x_B$, $y_B$]}\par
\medskip
\texttt{Determine which candidate location better matches the camera position of the}\par
\texttt{panoramic image.}\par
\medskip
\texttt{Compare the visual surroundings of each candidate in the satellite images with}\par
\texttt{what is visible in the panorama. Consider:}\par
\texttt{- Road patterns: intersections, lane markings, crosswalks, road width and curvature}\par
\texttt{- Buildings: color, size, shape, roof type, relative position to the road}\par
\texttt{- Vegetation: trees, their size and arrangement, proximity to the road}\par
\texttt{- Landmarks: parking lots, bridges, special structures, fences, walls}\par
\texttt{- Spatial layout: which side of the road features appear on, relative distances}\par
\texttt{between objects}\par
\texttt{- Depth-ring agreement (Images 5--6): the correct candidate's colored curve should}\par
\texttt{track the BLACK panorama curve more closely than the other candidate's does.}\par
\texttt{Compare the overall envelope and the azimuths of sharp transitions. If NEITHER}\par
\texttt{colored curve follows the black curve, the depth ring is uninformative for this}\par
\texttt{sample and you should rely on Images 1--4.}\par
\medskip
\texttt{Use Image 2 for global context, Images 3--4 for local detail around each}\par
\texttt{candidate, and Images 5--6 for cross-view geometric consistency. Focus on features}\par
\texttt{that clearly distinguish the two candidates. Prioritize strong, unambiguous}\par
\texttt{visual evidence over minor differences.}\par
\medskip
\texttt{Think step by step, then provide your final answer as \textbackslash boxed\{A\} or}\par
\texttt{\textbackslash boxed\{B\}.}\par
\end{minipage}}
\caption{Prompt for verification with depth-ring inputs.}
\label{fig:verifier-prompt-ring}
\end{figure}

\clearpage
\newlength{\verifierplainpromptwidth}
\begin{figure}[p]
\centering
\setlength{\fboxsep}{8pt}
\settowidth{\verifierplainpromptwidth}{\scriptsize\texttt{- Image 1: a ground-level $360^\circ$ panoramic view from the unknown camera location.}}
\fbox{%
\begin{minipage}{\verifierplainpromptwidth}
\scriptsize\raggedright
\textbf{User}\par\smallskip
\texttt{<image><image><image><image>}\par
\texttt{You are given four images:}\par
\texttt{- Image 1: a ground-level $360^\circ$ panoramic view from the unknown camera location.}\par
\texttt{- Image 2: the full satellite image with two candidate camera locations marked.}\par
\texttt{- Image 3: a $640\times640$ satellite crop centered on Candidate A (red marker).}\par
\texttt{- Image 4: a $640\times640$ satellite crop centered on Candidate B (blue marker).}\par
\medskip
\texttt{Candidate positions (normalized to [0, 1000]):}\par
\texttt{- A (Red): [$x_A$, $y_A$]}\par
\texttt{- B (Blue): [$x_B$, $y_B$]}\par
\medskip
\texttt{Determine which candidate location better matches the camera position of the}\par
\texttt{panoramic image.}\par
\medskip
\texttt{Compare the visual surroundings of each candidate in the satellite images with}\par
\texttt{what is visible in the panorama. Consider:}\par
\texttt{- Road patterns: intersections, lane markings, crosswalks, road width and curvature}\par
\texttt{- Buildings: color, size, shape, roof type, relative position to the road}\par
\texttt{- Vegetation: trees, their size and arrangement, proximity to the road}\par
\texttt{- Landmarks: parking lots, bridges, special structures, fences, walls}\par
\texttt{- Spatial layout: which side of the road features appear on, relative distances}\par
\texttt{between objects}\par
\medskip
\texttt{Use Image 2 for global context and Images 3--4 for local detail around each}\par
\texttt{candidate. Focus on features that clearly distinguish the two candidates.}\par
\texttt{Prioritize strong, unambiguous visual evidence over minor differences.}\par
\medskip
\texttt{Think step by step, then provide your final answer as \textbackslash boxed\{A\} or}\par
\texttt{\textbackslash boxed\{B\}.}\par
\end{minipage}}
\caption{Prompt for verification without depth-ring inputs.}
\label{fig:verifier-prompt-no-ring}
\end{figure}

\end{document}